%% file: main.tex
\documentclass[11pt,letterpaper]{article}
\usepackage[T1]{fontenc}
\usepackage[margin=1in]{geometry}
\usepackage[numbers,square,sort&compress]{natbib}
\usepackage{amsmath,amssymb,amsthm,graphicx,booktabs,placeins,longtable}
\usepackage{algorithm,algorithmic}
\usepackage{titletoc}

\let\amsEqref\eqref
\input{math_commands.tex}

\let\eqref\amsEqref
\renewcommand{\eps}{\varepsilon}

\usepackage{hyperref,url}
\hypersetup{colorlinks=true,linkcolor=blue,citecolor=blue,urlcolor=blue}
\newtheorem{theorem}{Theorem}
\newtheorem{proposition}[theorem]{Proposition}
\newtheorem{lemma}[theorem]{Lemma}

\theoremstyle{definition}

\theoremstyle{plain}
\newcommand{\OPT}{\operatorname{OPT}}
\newcommand{\err}{\operatorname{err}}
\title{Matching Under Metric Privacy\thanks{Code availability:
\url{https://github.com/hu-bryan/metric-private-matching}.}}
\author{%
  Hongyi B. Hu\textsuperscript{1}\quad
  Dung Nguyen\textsuperscript{2,3}\quad
  M. Amin Rahimian\textsuperscript{1}\\[0.5em]
  \small\textsuperscript{1}University of Pittsburgh\\
  \small\textsuperscript{2}Haverford College\\
  \small\textsuperscript{3}University of Texas at San Antonio
}
\date{November 20, 2026}
\begin{document}
\maketitle
\begin{abstract}
We study how to represent and randomize points in $d$-dimensional Euclidean space to obtain accurate bipartite matchings under metric privacy. Given $n$ private and $n$ public points, we analyze local mechanisms that randomize each private point or its distances to the public points. Using the public points' affine span, we construct a smaller representation that preserves all matching costs. When a trusted server receives the private points, we establish privacy and utility guarantees for Gumbel matching, which randomizes edge costs and releases only the assignment. Through upper bounds and worst-case constructions, we establish the tight rate $\Theta\!\left(n\min\{d,\log n\}/\varepsilon\right)$ for expected cost above the nonprivate optimum. Benchmark experiments support $d\approx\log n$ as a practical guide to the transition between directly randomizing point coordinates and using Gumbel matching. Extensions to squared Euclidean costs enable private nonparametric two-sample hypothesis testing. These results show how the application's trust model guides representation and dimension choices for better utility: under local privacy, low-dimensional point representations reduce excess cost, while under central privacy, distance representations achieve the optimal rate even in high dimension.
\end{abstract}

\section{Introduction}

Minimum cost matching has a long history in combinatorial optimization with well-established algorithms and numerous extensions \citep{burkard1999linear}. These algorithms are effective in applications such as allocation \citep{to2014framework,fahrbach2022edge}, data alignment and adaptation
\citep{courty2017transport,pmlr-v89-dai19b,shin2024otter}, and statistical
inference \citep{bai2022optimality,deb2023multivariate}.

Although there is an abundance of algorithms for matching, an observed output can be used together with any prior information to infer and expose the input, which is unacceptable in applications where the input data are sensitive. Differential privacy protects against this by limiting how much the output distribution can change from an arbitrary replacement of any one data point \citep{dwork2014algorithmic}. Metric privacy extends this definition by scaling this limit on distinguishability with the distance between inputs: small changes in the input require small changes in the output distribution \citep{chatzikokolakis2013broadening}.

In this work, we study matching minimizing the distance between two arbitrary sets $X, Y$, each with $n$ points in $\R^d$, and only those in $X$ are sensitive data. The goal is to release a matching $\sigma : i \mapsto j$ with small total matching cost $\sum_i \|\vx_i - \vy_{\sigma(i)}\|_2$ while satisfying metric privacy. This setting arises naturally when confidential target data are aligned with public source data in domain adaptation, or when workers are assigned to public tasks based on their private attributes.

Prior work on private matching differs from ours in two main ways.
First, the differential privacy literature studies matching with protection on the existence of node and edge \citep{dinitz2025differentially}, the tuple of edge weights \citep{sealfon2016shortest}, or provides joint differential privacy for allocations \citep{hsu2014private,hsu2016convex,huang2019scalable}. None of these works privatize the points' underlying numerical values.
Second, works in spatial crowdsourcing and vehicle assignment that study matching with metric privacy are restricted to two-dimensional locations and structured, finite metric spaces \citep{prorok2017assignment,tao2020differentially,wang2022graph,zhang2023canoe}. Thus, their methods do not extend to matching in general-dimensional Euclidean space.

We address this gap by examining how data in Euclidean space are represented and randomized and how this affects the output matching subject to metric privacy. A standard approach to sanitizing data in Euclidean space uses the Laplace mechanism, which perturbs each point by $\vx \mapsto \vx + \mathbf{Z}$, for $\mathbf{Z} \sim p(\vz) \propto \exp(-\eps\|\vz\|_2)$ \citep{hardt2010geometry,chatzikokolakis2013broadening}. This mechanism satisfies metric privacy and notably minimizes the reconstruction error $\E[\|\mathbf{Z}\|^2]$ \citep{koufogiannis2015optimality}. While data released this way are particularly suited to estimating aggregate statistics like centroids \citep{andres2013geo}, matchings computed from them can incur substantial excess cost over the non-private optimum.

We observe that public points $Y$ allow us to reduce the number of coordinates that require noise. Projecting the private points $X$ onto the affine span of $Y$ and retaining their distances from that span preserves every pair of distance between them. We can therefore add noise to fewer coordinates without discarding information needed for matching.

When a trusted platform computes the matching, it can keep the private points internal and release only the assignment that protects them under metric privacy. The privacy noise can therefore be designed to solely protect against inference from the matching. We adapt Gumbel matching \citep{mena2018learning}, based on the Gumbel-max trick, to add noise to $n^2$ edges rather than $n!$ matchings. To the best of our knowledge, we are the first to make this connection between Gumbel matching and metric privacy. In addition, we construct worst-case instances of $X,Y$ that show our mechanisms attain the lowest asymptotic excess cost possible under metric privacy.

An additional benefit is that all our mechanisms provide privacy and utility guarantees even when the cost is Euclidean distance squared, $\|\vx-\vy\|_2^2$. Minimizing this objective corresponds to minimizing mean squared error and maximizing the sum of dot products, both widely used in statistics and machine learning. We demonstrate the downstream value through nonparametric two-sample testing, where matching samples to a public reference grid constructs multivariate ranks \citep{deb2023multivariate}. Computing this matching privately extends the test to confidential samples.

To summarize our contributions:
\begin{itemize}
\item By privatizing individual data points, we show a matching could be produced with error $O\!\left(ns/\eps\right)$ for some $s \leq \min\{n, d\}$.
\item We determine that any $\eps D$-private matching mechanism achieves an error of $\Omega\!\left(n\min\{d,\log n\}/\eps\right)$ on some worst-case instance; this defines $d \approx \log n$ as the theoretical threshold between low- and high-dimensional regimes.
\item We obtain mechanisms that achieve upper bounds of $O\!\left(nd/\eps\right)$ and $O\!\left(n \log n/\eps\right)$ in the low- and high-dimensional regimes, respectively, making them optimal.
\item We empirically test the mechanisms on 16 benchmark datasets, examining how matching error varies with the privacy level, sample size, and data dimension. We further apply private matching to construct the multivariate ranks for nonparametric two-sample hypothesis testing. On the Sonar dataset, Gumbel matching preserves the most evidence for rejecting the null, obtaining a $p$-value below $0.05$ at $\eps=16$.

\end{itemize}

Appendix~\ref{app:related-work} reviews related work in metric privacy,
private optimization, geometric representations, and location-private assignment.

\section{Problem Setup and Privacy Preliminaries}
\label{sec:setup}

\paragraph{Matching.}
Let $X=(\vx_i)_{i=1}^n$ be the private dataset and
$Y=(\vy_j)_{j=1}^n$ the public dataset, whose data points are in $\sR^d$. Each permutation $\sigma \in S_n$ defines a matching between the private point $\vx_i$ and the public point $\vy_{\sigma(i)}$. We define the cost matrix $\mC=(C_{ij})$, utility $u(\sigma)$, and the optimum for the matching problem by
\begin{equation}\label{eq:matching-cost}
 \emC_{ij}=\|\vx_i-\vy_j\|_2,\qquad
 u(\sigma)=\sum_{i=1}^n\emC_{i,\sigma(i)},\qquad
 \OPT=\min_{\sigma\in S_n}u(\sigma).
\end{equation}
Throughout this paper, we analyze the excess matching cost $\err(\sigma)=u(\sigma)-\OPT$.



\paragraph{Metric privacy.}
Let the privacy metric between datasets be defined by
\begin{equation}\label{eq:dataset-metric}
 D(X,X')=\sum_{i=1}^n\|\vx_i-\vx_i'\|_2.
\end{equation}
Given a fixed and public $Y$, a mechanism $M_Y$ satisfies $\eps D$-privacy \citep{chatzikokolakis2013broadening} if
\begin{equation}\label{eq:dataset-privacy}
 \Pr[M_Y(X)\in A]
 \le \exp\!\bigl(\eps D(X,X')\bigr)\Pr[M_Y(X')\in A]
\end{equation}
for every $X,X'\in(\R^d)^n$ and every measurable event $A$. In our case, $M_Y$ is a probability distribution over the set of matchings.

Local metric privacy, a stronger notion of privacy, requires that every entry $i$ in the dataset independently privatizes its own data point by a mechanism $K_{i,Y}(\vx_i)$, with
\begin{equation}\label{eq:local-privacy}
 \Pr[K_{i,Y}(\vx)\in A]
 \le \exp\!\bigl(\eps\|\vx-\vx'\|_2\bigr)
       \Pr[K_{i,Y}(\vx')\in A]
\end{equation}
for every $\vx,\vx'\in\R^d$ and measurable $A$. In other words, $K_{i,Y}$ satisfies $\eps \|\cdot\|_2$-privacy on a singleton dataset. Subsequently, a mechanism $M_Y(X) = \left(K_{i,Y}(\vx_i)\right)_{i=1}^n$ defined by composing local metric-private mechanisms in parallel is $\eps D$-private.

Moreover, when two privacy mechanisms $M_1, M_2$ are composed sequentially, their privacy budgets add, satisfying $(\eps_1 + \eps_2)D$-privacy. Finally, any post-processing on the output from a private mechanism remains private. Their proofs follow standard arguments from differential privacy \citep{dwork2014algorithmic}.

Appendix~\ref{app:privacy-calibration} explains the connection to differential
privacy and how to interpret and calibrate $\eps$.

\paragraph{Laplace mechanism.}
One way to achieve local metric privacy is to add (generalized) Laplace noise.
For any $\eps > 0$ and norm
$\|\cdot\|$ on $\R^s$, the Laplace
mechanism maps $\vv\in\R^s$ to $\vv+\mathbf{Z}$, where the noise is independent
of the input and has density
\[
 p_{\mathbf{Z}}(\vz)\propto\exp(-\eps\|\vz\|).
\]
This mechanism satisfies $\eps\|\cdot\|$-privacy. When the norm is Euclidean, the noise decomposes as $\mathbf{Z}=R\bm{\Theta}$, where $R\sim\operatorname{Gamma}(s,\text{rate }\eps)$ and $\bm{\Theta}$ is independently uniform on the unit sphere $\sS^{s-1}$ \citep{awan2021structure}. For any chosen norm, the expected noise magnitude is $\E\|\mathbf{Z}\|=s/\eps$.

Several mechanisms in this work first process the private data points,
representing them by $h_Y(\vx_i)$, before adding noise. The Lipschitz
constant of $h_Y$ determines the noise scale needed to achieve
local $\eps$-metric privacy with respect to the original metric.
\begin{proposition}
\label{prop:lipschitz-calibration}
Suppose $h_Y:\R^d\to\R^s$ is $L$-Lipschitz for some $L>0$, meaning $\|h_Y(\vx)-h_Y(\vx')\|\le L\|\vx-\vx'\|_2$ for all $\vx,\vx'\in\R^d$. Then the Laplace mechanism applied to $h_Y(\vx)$ with ${\eps'=\eps/L}$ is $\eps\|\cdot\|_2$-private.
\end{proposition}

\subsection{Perturbing Points and Distances}
\label{sec:point-distance-perturbation}

The simplest approach is to add Laplace noise directly to each private point.
Each individual independently releases
$\widetilde\vx_i=\vx_i+\mathbf{Z}_i$, where $\mathbf{Z}_i$ has
density $p_{\mathbf{Z}_i}(\vz)\propto\exp(-\eps\|\vz\|_2)$.
We compute $\widehat{\emC}_{ij}=\|\widetilde\vx_i-\vy_j\|_2$ and return
\[
 \widehat\sigma\in\argmin_{\sigma\in S_n}
     \sum_{i=1}^n\widehat{\emC}_{i,\sigma(i)},
\]
using any minimum-cost matching algorithm.

\begin{theorem}
\label{thm:original-points}
The output matching satisfies $\eps D$-privacy and
\begin{equation}\label{eq:original-points-bound}
 \E[\err(\widehat\sigma)]\le\frac{2nd}{\eps}.
\end{equation}
\end{theorem}

This guarantee grows linearly with $d$, making it less useful in high
dimensions. However, matching only needs the edge weights
$\|\vx_i-\vy_j\|_2$, so we can replace each private point by its
distance vector
\[
 \Phi_Y(\vx)=\bigl(\|\vx-\vy_1\|_2,\ldots,\|\vx-\vy_n\|_2\bigr).
\]
This vector contains all costs in one row of $\mC$ and has $n$ coordinates,
regardless of $d$. For $p\in[1,\infty]$, individual $i$ independently releases
\[
 \widetilde\vc_i=\Phi_Y(\vx_i)+\mathbf{Z}_i,\qquad
 p_{\mathbf{Z}_i}(\vz)\propto
   \exp\!\left(-\frac{\eps}{n^{1/p}}\|\vz\|_p\right).
\]
We set
$\widehat{\emC}_{ij}=[\widetilde\vc_i]_j$ and compute a minimum-cost matching
for $\widehat{\mC}$.

\begin{theorem}
\label{thm:distance-vectors}
For every $p\in[1,\infty]$, the output matching satisfies $\eps D$-privacy and
\begin{equation}\label{eq:distance-vectors-p-bound}
 \E[\err(\widehat\sigma)]
 \le 2n\,\E\|\mathbf{Z}\|_p
 =\frac{2n^{2+1/p}}{\eps},
\end{equation}
which is minimized at $p=\infty$, giving
\begin{equation}\label{eq:distance-vectors-infinity-bound}
 \E[\err(\widehat\sigma)]\le\frac{2n^2}{\eps}.
\end{equation}
\end{theorem}

The proofs are in Appendix~\ref{app:matching-perturbation}.
The distance-vector bound no longer depends on $d$, but it grows quadratically
with $n$. Thus between the two methods, the tradeoff is either adding noise to $d$ coordinates or $n$ distances. However, the $n$ distances are computed from the same private point and public points whose affine span has dimension at most $d$. This suggests using that geometry to preserve the same distance information with fewer noisy coordinates.

\section{Matching Mechanisms}
\label{sec:mechanisms}

\subsection{Projected Representation}
\label{sec:projected-representation}

We use a representation by \citet{narayanan2019terminal} that preserves the distances between the public and private points. It retains each private point's projection
onto the public affine span and its distance from that span.
After translating both datasets so that
$\vy_1=0$, let
\[
 W=\operatorname{span}\{\vy_1,\ldots,\vy_n\},\qquad
 r=\dim W,\qquad s=\min\{r+1,d\}.
\]
Let $P_W$ and $P_{W^\perp}$ denote the orthogonal projections onto $W$ and
$W^\perp$. Writing projected vectors in a fixed public orthonormal basis of
$W$, define $f_Y:\R^d\to\R^s$ by
\begin{equation}\label{eq:projected-representation}
 f_Y(\vx)=
 \begin{cases}
  \bigl(P_W\vx,\|P_{W^\perp}\vx\|_2\bigr),&s<d,\\
  \vx,&s=d.
 \end{cases}
\end{equation}
The component perpendicular to $W$ contributes to each public-to-private
distance only through its magnitude, so just one additional coordinate suffices.
When this does not reduce dimension, we keep the original coordinates.
The representation preserves all matching costs:
\[
 \|f_Y(\vx)-f_Y(\vy_j)\|_2=\|\vx-\vy_j\|_2.
\]
Since $r\le n-1$, the representation uses at most $\min\{n,d\}$ coordinates.

Each individual independently releases
\[
 \widetilde\vx_i=f_Y(\vx_i)+\mathbf{Z}_i,\qquad
 p_{\mathbf{Z}_i}(\vz)\propto\exp(-\eps\|\vz\|_2),
\]
with noise in $\R^s$. We compute
$\widehat{\emC}_{ij}=\|\widetilde\vx_i-f_Y(\vy_j)\|_2$ and return a
minimum-cost matching for $\widehat{\mC}$.

\begin{theorem}
\label{thm:projected-points}
The output matching satisfies $\eps D$-privacy and
\begin{equation}\label{eq:projected-points-bound}
 \E[\err(\widehat\sigma)]\le\frac{2ns}{\eps}
 \le\frac{2n\min\{n,d\}}{\eps}.
\end{equation}
\end{theorem}

The proof is in Appendix~\ref{app:projected-points-proof}.

\subsection{Gumbel Matching}
\label{sec:gumbel-matching}

When the platform computing the matching is trustworthy, local metric privacy can be exchanged for better utility. We can randomize the assignment directly
instead of releasing a noisy representation of each private point.
We use Gumbel matching, which adds independent Gumbel noise to the edge
scores and computes a maximum-weight matching \citep{mena2018learning}.
The standard Gumbel distribution, $\operatorname{Gumbel}(0,1)$, has
cumulative distribution function
\[
 \Pr[G\le t]=\exp(-e^{-t}),\qquad t\in\R.
\]
For our Euclidean costs, draw independent $G_{ij}\sim\operatorname{Gumbel}(0,1)$
and return
\begin{equation}\label{eq:gumbel-matching}
 \widehat\sigma\in\argmax_{\sigma\in S_n}
 \sum_{i=1}^n\left(-\emC_{i,\sigma(i)}
                    +\frac2\eps G_{i,\sigma(i)}\right).
\end{equation}

\begin{theorem}
\label{thm:gumbel-matching}
For every $X,Y\in(\R^d)^n$ and $\eps>0$, Gumbel matching satisfies $\eps D$-privacy and
\begin{equation}\label{eq:gumbel-matching-bound}
 \E[\err(\widehat\sigma)]\le\frac{2n\log n}{\eps}.
\end{equation}
\end{theorem}

The proof is in Appendix~\ref{app:gumbel-matching-proof}.

\section{A Privacy Lower Bound}
\label{sec:privacy-lower-bound}

The preceding guarantees leave open whether another mechanism can improve
the dependence on $n$ and $d$.
We now bound the error required by metric privacy.

\begin{theorem}
\label{thm:privacy-lower-bound}
For every $n\ge2$, $d\ge1$, and $\eps>0$, there exists a public dataset
$Y\in(\R^d)^n$ such that every $\eps D$-private matching mechanism $M_Y$
has a private input $X\in(\R^d)^n$ for which its output
$\widehat\sigma=M_Y(X)$ satisfies
\begin{equation}\label{eq:privacy-lower-bound}
 \E[\err(\widehat\sigma)]=\Omega\!\left(\frac{n\min\{d,\log n\}}{\eps}\right).
\end{equation}
\end{theorem}

Our proof follows a packing approach (e.g. see \citet{hardt2010geometry}) to obtain privacy lower bounds: with points in $Y$ $\Omega(h)$-apart, we construct a family of $\{X^\pi\}$, each with optimal matching $\pi$, such that
\[
\err(\sigma)\gtrsim h\,\#\{i:\sigma(i)\ne\pi(i)\}.
\]
We can show some $X^\pi$ achieves $\Omega(n)$ such disagreements by showing that a uniformly random one does in expectation. This can be shown via a constant lower bound on $\Pr[\sigma(i) \ne \pi(i)]$ across $\Omega(n)$ indices $i \in [n]$. Across $X^\pi$, $\pi(i)$ takes $k$ distinct values, with $\log k=\Theta(\min\{d,\log n\})$. Inputs with $\pi(i)=a$ can be paired with inputs with $\pi'(i)=b$ by exchanging $a,b$ in the optimal matching, while keeping $D(X^\pi,X^{\pi'})=O(h)$. Metric privacy therefore bounds the probability ratios for any released partner between these cases by \(e^{O(\eps h)}\), including after averaging over the remaining choices of \(\pi\). Taking \(h\) as a sufficiently small constant multiple of \(\log k/\eps\), these bounds and probability normalization force the desired probability lower bound.

The complete construction and proof are in
Appendix~\ref{app:privacy-lower-bound-proof}.

Choosing original-point perturbation when $d\le\log n$ and Gumbel matching
otherwise gives
\[
 \E[\err(\widehat\sigma)]\le\frac{2n\min\{d,\log n\}}{\eps}.
\]
The choice depends only on the public $n,d$ and preserves $\eps D$-privacy. Notably, the $d \approx \log n$ threshold separates low- vs.\ high-dimensional regimes. 

\section{Extensions}
\label{sec:extensions}

\paragraph{Squared Euclidean Matching.}
\label{sec:squared-matching}

The same four mechanisms also apply when the matching cost is squared
Euclidean distance:
\[
 \emC_{ij}=\|\vx_i-\vy_j\|_2^2.
\]
The total cost expands as
\begin{equation}\label{eq:squared-matching-objective}
\begin{aligned}
 u(\sigma)&=\sum_i\emC_{i,\sigma(i)}
            =\sum_i\|\vx_i-\vy_{\sigma(i)}\|_2^2\\
 &=\sum_i\|\vx_i\|_2^2+\sum_j\|\vy_j\|_2^2
       -2\sum_i\langle\vx_i,\vy_{\sigma(i)}\rangle.
\end{aligned}
\end{equation}
Because a bijection uses every point once, the first two terms do not depend on the matching. Minimizing the objective
is therefore equivalent to maximizing the total matched dot product:
\[
 \argmin_{\sigma\in S_n}u(\sigma)
 =\argmax_{\sigma\in S_n}\sum_i\langle\vx_i,\vy_{\sigma(i)}\rangle.
\]
The mean squared error and sum of dot products are ubiquitous in statistics and machine
learning. Appendix~\ref{app:squared-matching} gives their expected excess-cost upper
bounds under this objective.

\paragraph{Matching When Both Datasets Are Private.}
\label{sec:both-private-matching}

When both $X$ and $Y$ are sensitive, privacy must protect changes to
either dataset. Appendix~\ref{app:both-private-matching} gives the
corresponding mechanisms and upper bounds for Euclidean matching.

\section{Numerical Experiments}
\label{sec:exp-numerical}

We examine how the theoretical dependence on privacy, dimension, and
sample size appears on benchmark data. The experiments address three
questions: how utility improves as the privacy budget increases, how the
choice between point perturbation and direct matching changes with
dimension, and how increasing the number of matching partners affects
error. We use real-world and synthetic
benchmarks with varied geometries, so that these comparisons extend
beyond any single distribution.

\subsection{Experimental Setup}
\label{sec:exp-numerical-setup}

We compare \emph{Original points}, \emph{Projected representation},
\emph{Distance vectors}, and \emph{Gumbel} under the Euclidean privacy
metric used in the theoretical analysis. The privacy-budget sweeps use
both Euclidean costs $C_{ij}=\|\vx_i-\vy_j\|_2$ and squared Euclidean costs,
$C_{ij}=\|\vx_i-\vy_j\|_2^2$; the dimension and sample-size sweeps use
Euclidean costs. Appendix~\ref{app:exp-mechanisms} gives the calibration
for each objective.

Our main comparison uses sixteen datasets. Fifteen come from the
versioned \emph{Benchmark Suite for Clustering Algorithms}
\citep{gagolewski2022framework,gagolewski2026suite}, spanning synthetic
shapes, biological and physical measurements, and images, with dimensions
from 2 to 784. We add Banknote Authentication
\citep{lohweg2012banknote}: its four features and 1,372 records allow a
sample-size sweep on both sides of $\log n=4$ using disjoint samples.
Table~\ref{tab:exp-datasets} gives each dataset's original source and
available size. Class labels are excluded from the features and play no
role in sampling or matching.

\begin{table}[t]
 \centering\small
 \caption{Experimental configurations. The dimension and sample-size
 sweeps vary one quantity while holding the other fixed.}
 \label{tab:exp-configurations}
 \setlength{\tabcolsep}{5pt}
 \begin{tabular}{llll}
  \toprule
  Experiment & Dataset & $n$ per side & Budget and objective\\
  \midrule
  Privacy sweep & All sixteen & 64 & $\eps=0.25,0.50,\ldots,16$; both costs\\
  Fixed-budget bars & All sixteen & 64 & $\eps=8$; both costs\\
  Dimension sweep & Fashion-MNIST & 800 & $\eps=16$; Euclidean\\
  Sample-size sweep & Banknote & $20,40,\ldots,200$ & $\eps=8$; Euclidean\\
  \bottomrule
 \end{tabular}
\end{table}

\paragraph{Feature and distance scales.}
Tabular datasets can mix features with different units and numerical
ranges. We standardize their varying numerical features by subtracting
their means and dividing by their standard deviations, while mapping
binary indicators to zero and one before a final common rescaling.
For example, an Ecoli record combines numerical sequence scores with
binary indicators; we standardize the scores and retain each indicator
as one coordinate. This follows standard input normalization in machine
learning \citep{goodfellow2016deep} and sets the relative feature weights
in the matching distance. Iris, the synthetic datasets, and the image
datasets have consistent units across coordinates, so we retain their
relative feature scales.

After this feature treatment, every actual matching instance is centered
and divided by its pooled root-mean-square (RMS) radius, giving RMS radius
one. Privacy guarantees apply to these normalized inputs, with preprocessing
held fixed.

\paragraph{Dimension and sample-size sweeps.}
To examine how error grows with dimension, we fix $n=800$ and project
Fashion-MNIST images onto nested random subspaces of decreasing
dimension, starting from all 784 coordinates. To examine how error grows
with sample size, we fix $d=4$ and take nested Banknote samples with
$n=20,40,\ldots,200$ points per side. Both sweeps repeat the centering
and RMS rescaling at every setting, keeping the feature weights fixed.
Thus changes in error reflect the varying dimension or sample size at
a common overall distance scale and privacy budget.
Appendix~\ref{app:exp-normalization}
gives the exact transformations and their relation to the privacy
calibration in Appendix~\ref{app:privacy-calibration}.

\paragraph{Evaluation.}
In every experiment, we draw twenty independent collections of records,
sampling without replacement within each collection. Each collection
provides five partitions, each with disjoint public and private sides, and, in the dimension
sweep, five random projection chains. We solve the assignment problem
exactly after perturbation and evaluate excess cost on the true costs.
We compute the reference for a uniformly random matching exactly as
$\mathrm{RAND}=n^{-1}\sum_{i,j}C_{ij}$.
Both the optimum and the random reference are evaluated separately for
each instance and objective.

We average 50 independent mechanism runs for each fixed matching
instance, then average over partitions and projection chains within
each collection. The reported mean averages these twenty collection
means. Appendix~\ref{app:exp-numerical-values} reports their confidence
intervals, with the calculation detailed in
Appendix~\ref{app:exp-mechanisms}.

\subsection{Results}
\label{sec:exp-numerical-results}

\begin{figure}[h]
 \centering
 \includegraphics[width=\linewidth]{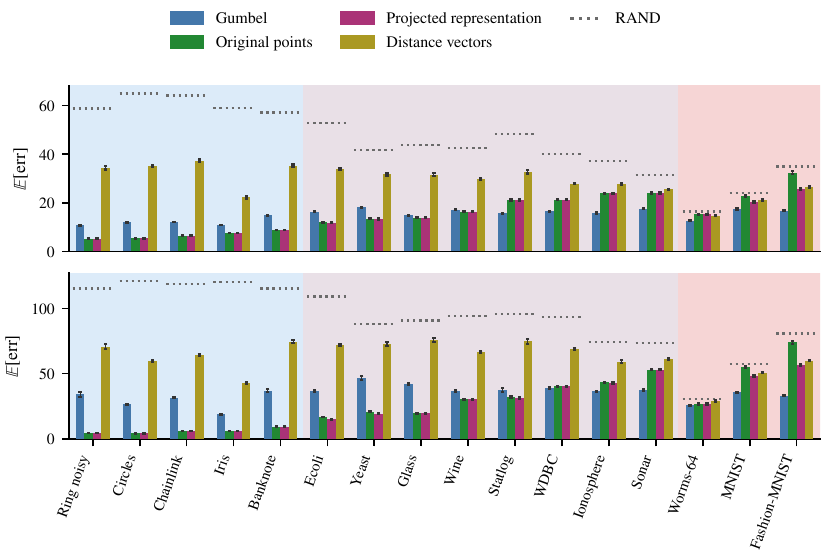}
 \caption{Excess matching cost at $\eps=8$, $n=64$, for Euclidean
 (top) and squared Euclidean (bottom) objectives. Blue blocks indicate
 $d\le\log n$, rose blocks $d\ge n$, and their equal blend
 $\log n<d<n$. Within each dataset, bars follow legend order. Dotted segments give
 the random reference for that dataset and objective.}
 \label{fig:exp-benchmark-bars}
\end{figure}

\begin{figure}[h]
 \centering
 \includegraphics[width=\linewidth]{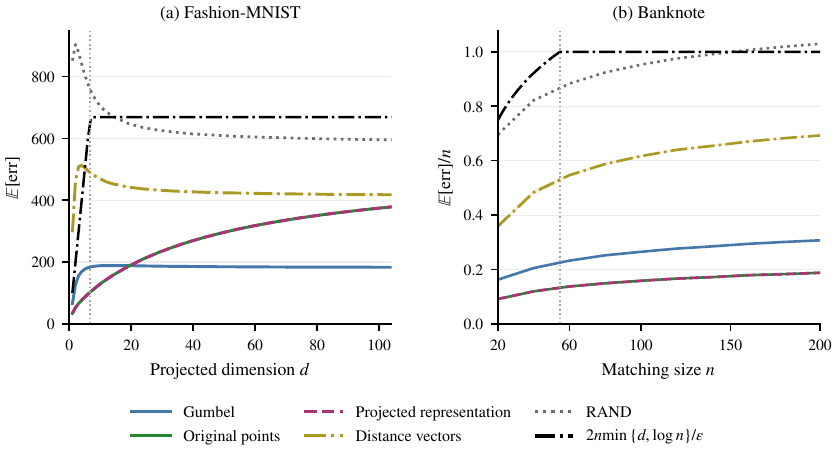}
 \caption{Dimension and sample-size sweeps with RMS rescaling.
 (a) Fashion-MNIST: $n=800$, $\eps=16$, total excess cost.
 (b) Banknote: $d=4$, $\eps=8$, excess cost per pair.
 Gray dotted curves give the corresponding random references. Black dash-dot
 curves give $2n\min\{d,\log n\}/\eps$ in (a) and
 $2\min\{d,\log n\}/\eps$ in (b). Vertical dotted lines mark
 $d=\log(800) \approx 6.68$ and $n=e^4\approx 54.60$, respectively.}
 \label{fig:exp-dimension-and-size}
\end{figure}

\paragraph{Privacy-budget sweeps and benchmark comparisons.}
Increasing $\eps$ generally reduces excess matching cost
(Appendix~\ref{app:exp-epsilon-sweeps}).
Figure~\ref{fig:exp-benchmark-bars} shows that the mechanisms' relative
performance varies across datasets and objectives.
For Euclidean costs, the point methods give 16.43 on Wine against
Gumbel's 17.27, while WDBC favors Gumbel, 16.48 against 21.26.
For squared costs, Projected representation gives 14.92 on Ecoli against
Gumbel's 36.54, while WDBC again favors Gumbel.

\paragraph{Dimension sweep.}
In Figure~\ref{fig:exp-dimension-and-size}(a), the two point methods
coincide and their excess cost increases with dimension.
Gumbel overtakes them between $d=19$ and 24.
Both Gumbel and Distance vectors approach plateaus, with Gumbel
remaining lower.
The combined theoretical guarantee lies below the mean random
reference through the measured dimension $d=14$.
Table~\ref{tab:exp-fashion-values} gives the full numerical comparison.

\paragraph{Sample-size sweep.}
In Figure~\ref{fig:exp-dimension-and-size}(b), the point methods have
lower excess cost per pair than Gumbel throughout.
Their curves and Gumbel's gradually flatten, whereas Distance vectors
continues to rise more noticeably.
Table~\ref{tab:exp-banknote-values} gives the numerical values.

\subsection{Interpretation}
\label{sec:exp-numerical-interpretation}

Gumbel's advantage shows that improving the representation is not
the only way to improve private matching. Gumbel and Distance vectors
perturb the same matching costs, but impose privacy on different
outputs. Distance vectors protects the entire report, whereas Gumbel
keeps the perturbed costs inside a trusted server and releases only
the assignment. The latter can therefore use noise calibrated to
protect the matching without making every reported distance private.

This distinction gives a logarithmic dependence on the number of
candidates rather than a dependence on the ambient dimension.
The $\log n$ term comes from the expected maximum of the $n$
Gumbels in each row. Together with the lower bound, this establishes
that Gumbel attains the optimal excess-cost order in high dimension.
Its empirical advantage therefore accompanies a theoretical
guarantee that applies beyond the evaluated datasets.

The dimension and sample-size sweeps examine the comparison with
point perturbation from two directions. Decreasing $d$ or increasing
$n$ moves toward $d\le\log n$, where Original points already attains
the optimal order under local metric privacy. Projected representation
can further reduce the number of coordinates requiring noise when
the public affine span permits it. Together, the theoretical bounds
and experimental comparisons support using $d\approx\log n$ as a
practical guide for gauging when the advantage shifts from point
perturbation toward Gumbel matching.

\section{Nonparametric Two-Sample Hypothesis Testing}
\label{sec:exp-rank-testing}

\citet{deb2023multivariate} develop a distribution-free nonparametric
two-sample test based on multivariate ranks. Their empirical rank map
matches the pooled observations to a fixed reference grid by minimizing
the total squared Euclidean distance. The resulting rank-energy
statistic compares the two empirical rank distributions.

When the observations are confidential, privatizing the rank assignment
provides a natural way to perform this test under metric privacy. The
pooled observations form the private side of the matching, and the equally
sized reference grid forms the public side. We describe the two-sample setting
below, then evaluate the private test on Sonar.

\subsection{The Two-Sample Rank-Energy Test}
\label{sec:exp-rank-procedure}

Let $X=(\vx_i)_{i=1}^N$ pool two independent samples in $\R^d$, with
$\vx_1,\ldots,\vx_a$ drawn from $F$ and $\vx_{a+1},\ldots,\vx_N$ from $G$,
where $N=a+b$. The cohort labels are public. We test
\[
 H_0:F=G\qquad\text{versus}\qquad H_1:F\ne G.
\]
Fix $N$ distinct public points $Y=(\vy_j)_{j=1}^N\subset[0,1]^d$,
independently of the observations. The empirical multivariate rank of
$\vx_i$ is the grid point $\vy_{\sigma^\star(i)}$ assigned by
\begin{equation}
\label{eq:exp-rank-assignment}
 \sigma^\star\in\argmin_{\sigma\in S_N}
       \sum_{i=1}^N\|\vx_i-\vy_{\sigma(i)}\|_2^2.
\end{equation}
For any assignment $\sigma$, define the rank-energy statistic by
\begin{equation}
\label{eq:exp-rank-energy}
\begin{aligned}
 \mathcal E(\sigma)={}&
 \frac{2}{ab}\sum_{i=1}^a\sum_{j=a+1}^N
                \|\vy_{\sigma(i)}-\vy_{\sigma(j)}\|_2\\
 &-\frac{1}{a^2}\sum_{i,j=1}^a
                \|\vy_{\sigma(i)}-\vy_{\sigma(j)}\|_2\\
 &-\frac{1}{b^2}\sum_{i,j=a+1}^N
                \|\vy_{\sigma(i)}-\vy_{\sigma(j)}\|_2,
 \qquad T(\sigma)=\frac{ab}{N}\mathcal E(\sigma).
\end{aligned}
\end{equation}
The energy statistic uses unsquared Euclidean distances to measure
separation between the empirical rank distributions. Large $T$ provides
evidence against $H_0$.

We replace $\sigma^\star$ with an $\eps D$-private assignment
$\widehat\sigma$ and compute $T(\widehat\sigma)$ from its ranks.
Under $H_0$, the exact-rank statistic has a \emph{distribution-free null
distribution}: its law depends only on $Y,a,b$, independently of the
unknown common distribution of the observations
\citep[Section~4.2]{deb2023multivariate}. The same property holds after
privatization when the matching rule ignores cohort labels and treats
observation rows symmetrically. We calculate an upper-tail $p$-value from this null
distribution and reject at $p\le0.05$.
Appendix~\ref{app:exp-rank-details} describes how we approximate the null
distribution and calculate the $p$-value. The statistic and decision
are post-processing of the private assignment.

\subsection{Experimental Design on Sonar}
\label{sec:exp-sonar-design}

Following the application in \citet{deb2023multivariate},
we test whether Sonar measurements reflected from metal cylinders and
rocks have the same distribution. The UCI Sonar dataset contains $a=111$ metal-cylinder and $b=97$
rock observations, each with $d=60$ signal-energy coordinates in $[0,1]$
\citep{sejnowski1988sonar}.
We retain all $N=208$ observations in their original units, without
standardization or feature selection. Although Sonar is a public benchmark,
we treat every measurement vector as confidential; cohort labels and the
roster are public.

We implemented the squared-cost rank matching and rank-energy test above,
using a fixed reference grid consisting of the first 208 Halton points in
60 dimensions. The same grid is used for every method and budget. We
compare \emph{Original points}, \emph{Projected representation},
\emph{Distance vectors}, and \emph{Gumbel}, with squared Euclidean
matching costs.

For each mechanism and $\eps\in\{1,2,4,8,16\}$, we evaluate
1,000 independent runs on this fixed dataset. We calculate $p$-values
using 9,999 independent draws from the null distribution, generated
once and shared across all methods and runs. We report the mean
$p$-value over the 1,000 runs, with the exact non-private ranks as a baseline.

\subsection{Results and Interpretation}
\label{sec:exp-sonar-results}

\begin{figure}[t]
 \centering
 \includegraphics[width=0.65\linewidth]{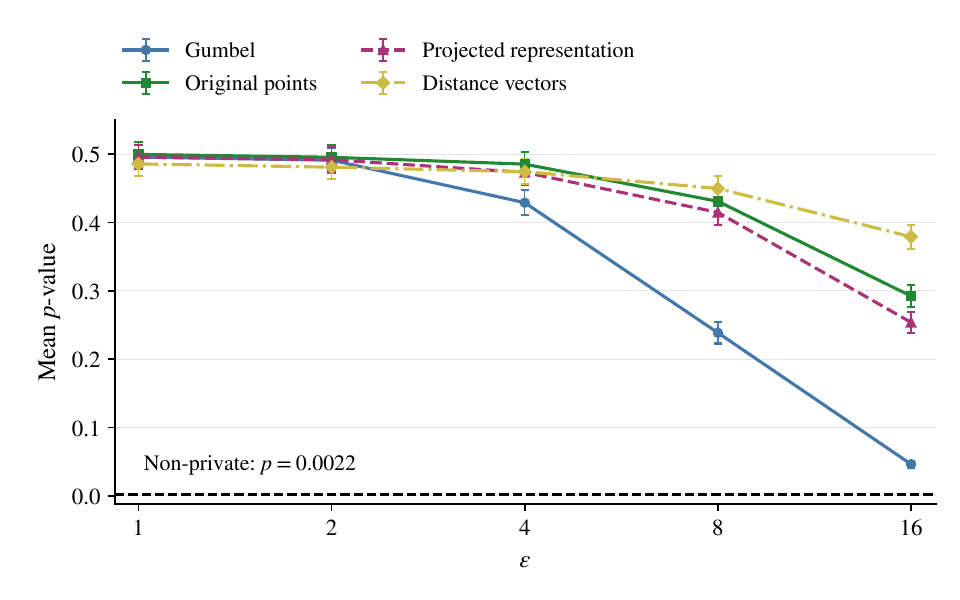}
 \caption{Mean rank-energy $p$-values on Sonar, using the same dataset and simulated null distribution.
 Error bars are pointwise 95\% confidence intervals for the means.}
 \label{fig:exp-sonar-mean-pvalue}
\end{figure}

The exact non-private matching in our experiment gives $T=7.064$ and
$p=0.0022$, rejecting $H_0$ at the 5\% level.
Figure~\ref{fig:exp-sonar-mean-pvalue} shows the mean $p$-values produced
by private matching, with this non-private result as a fixed baseline.

At $\eps=1, 2$, all four methods give mean $p$-values close to
0.5, showing that strong privacy obscures the difference between the two
groups. As $\eps$ increases, the private test becomes more informative.
Gumbel's mean $p$-value falls from 0.4290 at $\eps=4$ to 0.2386 at
$\eps=8$ and 0.0463 at $\eps=16$. At $\eps=16$, Projected
representation gives 0.2539, Original points 0.2924, and Distance vectors
0.3791. Allowing a larger privacy budget therefore helps recover evidence
of the distributional difference, with Gumbel coming closest to the
non-private baseline on this dataset.
Appendix~\ref{app:privacy-dataset-calibration} interprets these privacy
budgets in terms of an attacker's ability to distinguish Sonar data points.

\section{Discussion}

The application's trust model should guide the representation and dimension
chosen for sensitive data. Under local metric privacy, low-dimensional embeddings
reduce the number of coordinates requiring noise; the public affine span
permits this reduction while preserving all matching costs. With a trusted
server, Gumbel matching can retain high-dimensional distance information and
attain the optimal excess-cost rate. Thus, $d\approx\log n$ offers guidance
for both mechanism choice and embedding design. The optimal high-dimensional
local rate remains open.

Gumbel matching also connects metric privacy to learning latent permutations.
\citet{mena2018learning} introduce Gumbel-Sinkhorn as a differentiable
relaxation for learning with random assignments. Our analysis identifies a
complementary property of discrete Gumbel matching: controlling changes in the edge costs protects the entire assignment despite its global
constraints. This suggests studying Gumbel constructions for other
metric-private optimization problems and learning tasks involving latent
correspondences. Extending these guarantees to learned scores and released
soft assignments would bring metric privacy closer to differentiable learning.

The hypothesis-testing application shows how private matching can preserve
information useful beyond the assignment itself. Relating excess matching
cost to statistical power would help connect our optimization guarantees to
downstream inference. Online matching provides another direction,
particularly for allocation problems where workers or tasks arrive
sequentially and assignments must be made immediately. Studying how privacy
noise interacts with uncertain arrivals and irrevocable decisions could
extend the representation and mechanism choices developed here to dynamic
allocation.

\subsection*{AI use statement}
Generative AI tools were used throughout the research and preparation of this paper. Their uses included identifying and summarizing relevant literature; refining research ideas; developing mathematical arguments; drafting, checking, and revising proofs; designing experiments; implementing, debugging, and running experimental code; processing data; analyzing results; preparing figures, tables, and reproducibility materials; and editing manuscript wording, grammar, and clarity. 

These activities were conducted under the authors’ direction. The authors take full responsibility for the final manuscript, including the correctness and originality of its theoretical claims, proofs, experimental methods, results, and all other AI-assisted contributions.

\clearpage
\bibliographystyle{plainnat}
\bibliography{references}

\clearpage
\appendix

\startcontents[appendix]
\section*{Appendix Contents}
\printcontents[appendix]{}{1}[2]{}
\clearpage

\section{Related Work}
\label{app:related-work}

\subsection{Metric Privacy}

Metric privacy bounds distinguishability between inputs according to their distance, protecting precise values while allowing disclosure of coarser information.
\citet{chatzikokolakis2013broadening} propose the general framework, while \citet{andres2013geo} develop geo-indistinguishability for location-based systems.
For a broader account of the development of metric privacy, its mechanism families, and its applications, we refer readers to the systematization of knowledge by \citet{xie2026sok}.

Subsequent work investigates how mechanisms optimize the tradeoff between privacy and utility.
\citet{bordenabe2014optimal} design location-release mechanisms subject to geo-indistinguishability, and \citet{imola2022balancing} address the computational challenges of constructing useful metric-private mechanisms at scale.
More generally, geometric approaches relate privacy--utility tradeoffs to the geometry of the released statistics and the input metric \citep{hardt2010geometry,boedihardjo2024metric}.
The choice of input metric also determines what information is protected: for example, \citet{imola2024earth} use the Earth-Mover's distance to formulate user-level metric privacy for collections of records.
This use of transport as an input privacy metric is distinct from minimizing a transport or matching objective.
\citet{alvim2018local} study the utility of local metric privacy;
\citet{koufogiannis2015optimality} analyze additive mechanisms for numerical
reconstruction under Lipschitz privacy. The latter criterion differs from
the assignment loss studied here.

\subsection{Private Optimization and Matching}

Private matching belongs to the broader study of optimization with sensitive problem data.
\citet{hsu2014linear} distinguish classes of private linear programs according to whether sensitive information appears in the objective or constraints, providing algorithms and impossibility results for these different formulations.
Recent work further studies private linear-program feasibility and the tradeoff between constraint satisfaction and optimization accuracy \citep{ene2025linear,benvenuti2026linear}.
These distinctions are directly relevant to assignment: its linear-program formulation has a fixed feasible region when the matching constraints are public, while private locations determine the objective coefficients.
The geometric dependence between these coefficients supplies additional structure beyond a general private objective vector.

Another line of work considers optimization in which different agents receive different parts of the solution.
\citet{hsu2014private} study private matchings and allocations under joint differential privacy, and \citet{hsu2016convex} extend this perspective to a broad family of convex programs with shared resource constraints.
Joint privacy protects an agent's information against the outputs received by other agents, while allowing that agent's own allocation to depend on its input.
\citet{danassis2022distributed} develop decentralized resource allocation under a piecewise local privacy formulation.
Resource-allocation privacy also arises through observable contention for shared resources, as studied by \citet{chen2023allocation}.
Together, these works highlight the importance of specifying both the sensitive information and the observer's access to the solution.

Matching-specific results address several further privacy models.
\citet{dinitz2025differentially} study matching under edge and node privacy, distinguishing explicitly released matchings from implicit representations.
For minimum-weight perfect matching, \citet[Appendix B.2]{sealfon2016shortest} gives algorithms and lower bounds when the topology is public and edge weights are private under an $\ell_1$ neighboring relation.
\citet{imola2026lower} strengthen reconstruction-based lower bounds in this setting.
These results provide direct optimization precedents, although arbitrary changes to edge weights differ from the coordinated changes to matching costs induced by moving a private point.
\citet{anandan2017secure} combine secure bipartite-matching computation with differential privacy for the numerical matching cost, illustrating the distinction between protecting an optimum value and protecting a released assignment.

Private optimal transport is closely related because, for two equally weighted point sets of the same size, unregularized discrete transport admits an optimal permutation coupling; its objective equals the corresponding minimum matching cost divided by the number of points.
\citet{letien2019transport} use private projected representations and reconstructed transport costs for domain adaptation.
\citet{wang2024sinkhorn} develop a private Sinkhorn procedure for entropically regularized transport, together with a further randomized transport-cost release.
These methods connect private geometric representations to constrained allocation, while their privacy models and, in the latter case, regularized objective differ from matching under metric privacy with unsquared Euclidean costs.

\subsection{Private Geometric Representations and Distance Queries}

Private representations can preserve information needed for a downstream task without accurately reconstructing every original coordinate.
\citet{kenthapadi2013privacy} combine Johnson--Lindenstrauss projection with Gaussian perturbation and correct the resulting bias in squared-distance estimates.
\citet{huang2014metric} exploit metric structure for private distance-query release.
More recently, \citet{backurs2024similarities} develop private data structures for aggregate similarities and distances to a dataset.
Aggregate query answers and point-specific assignment costs serve different purposes, but these works establish a common connection between geometric structure, private representations, and downstream utility.
Private projected representations have also been used for Euclidean minimum spanning trees \citep{zou2026mst}, providing an additional example of geometry informing private optimization.

Public reference points offer a further source of structure.
Terminal dimensionality reduction preserves distances from designated terminals to other points.
The exact projection-plus-orthogonal-norm representation used in this work appears in \citet[Section 1.1]{narayanan2019terminal}.
Both dimensionality reduction and reconstruction of distances from private sketches therefore have established precedents.
Our projected-representation mechanism adds calibrated noise to this distance-preserving representation and bounds the resulting excess matching cost.

Classical random assignment studies optimal matching costs under random edge weights.
\citet{linusson2004parisi} prove the exact expected optimum for a square
matrix of independent exponential costs. In private optimization,
\citet{hardt2010geometry} use the geometry of norm balls to construct
noise distributions and establish privacy--utility bounds.

\subsection{Location-Private Assignment and Spatial Crowdsourcing}

Spatial crowdsourcing provides an extensive application literature on assignment with sensitive locations.
Early work uses private spatial decompositions to support task dissemination while protecting worker populations \citep{to2014framework}.
\citet{prorok2017assignment} study location-private vehicle assignment for mobility-on-demand systems.
Tree-based representations support online task assignment \citep{tao2020differentially}; quadtree generalization supports personalized privacy \citep{lv2022personalized}; and graph-based approaches support online allocation after location perturbation \citep{wang2022graph}.
\citet{li2021batch} consider batch-based task assignment with an untrusted server.
\citet{du2023dynamic} incorporate changing privacy choices and compensation into task assignment.
\citet{zhang2023canoe} propose CANOE, which groups workers' preferred tasks and collects distances from their true locations to perturbed group representatives, providing a closely related approach to supporting assignment through obfuscated distance information.

These formulations differ in arrival assumptions, capacities, disclosure, and utility objectives.
For example, \citet{xu2026ridesharing} study online ridesharing with known-distribution arrivals and driver matching budgets, considering matching count, profit, and pickup distance.
The privacy metric also matters: the mechanism of \citet{lv2022personalized} uses differences between numbered quadtree nodes at a selected level, rather than continuous Euclidean displacement.
This literature motivates private assignment and provides application-specific mechanisms.
The present formulation isolates the optimization problem of releasing a minimum-total-distance perfect matching, with public assignment constraints and an additive utility guarantee that does not require bounded or distributionally specified private locations.

\clearpage
\section{Differential Privacy and Privacy Calibration}
\label{app:privacy-interpretation}

\subsection{Differential Privacy and the Meaning of $\eps$}
\label{app:privacy-calibration}

In differential privacy, two datasets of the same size are neighbors if
one record differs, regardless of the magnitude of that difference. A
mechanism $M_Y$ satisfies $\eps$-differential privacy if, for every such
pair $X,X'$ and every measurable event $A$,
\[
 \Pr[M_Y(X)\in A]\le e^\eps\Pr[M_Y(X')\in A].
\]
This is the replacement-neighbor definition of differential privacy
\citep{dwork2014algorithmic}. Define the Hamming distance by
\[
 d_H(X,X')=\sum_{i=1}^n\1_{\{\vx_i\ne\vx_i'\}}.
\]
Substituting $d_H$ for $D$ in \eqref{eq:dataset-privacy} recovers this
notion: neighbors have Hamming distance one, and replacing differing
records one at a time gives the converse implication for arbitrary pairs
\citep{chatzikokolakis2013broadening}.

With our Euclidean metric $D$, suppose instead that neighbors differ in
one record by distance at most one. Then $D(X,X')\le1$, so metric privacy
implies $\eps$-differential privacy for this neighborhood relation.
More generally, choose a distance $r>0$ and a tolerance factor $c>1$:
for inputs at distance at most $r$, the probability of any output event
should change by at most a factor $c$. Following the interpretation of metric privacy by \citet{andres2013geo}, set
\[
 \eps=\frac{\log c}{r}.
\]
Indeed, whenever $D(X,X')\le r$,
\[
 \Pr[M_Y(X)\in A]
 \le e^{\eps r}\Pr[M_Y(X')\in A]
 =c\Pr[M_Y(X')\in A].
\]
For a change to one
record, $r$ is the Euclidean displacement of that point.

The radius $r$ is measured in the units of the data, whether physical
units or distances in a latent representation, and $\eps$ has inverse
units so that $\eps r$ is dimensionless.
Changing from meters to centimeters, for example, multiplies numerical
distances by $100$ and divides $\eps$ by $100$, preserving the same
privacy guarantee. Likewise, calibration in a latent space depends on
the scale of its coordinates.

\subsection{Calibration on the Experimental Dataset}
\label{app:privacy-dataset-calibration}

Consider an attacker who knows all private data points and their identities,
except which of two individuals has which of two known points
$\vx,\vx'\in\R^d$. The attacker considers two equally likely datasets
$X_0,X_1$ that differ only by exchanging these points. After observing
one matching, including the assignments to the individuals, the attacker
guesses which dataset was used. Observing the non-private matching has $100\%$ attack accuracy, while a uniformly random one has $50\%$. 

\begin{figure}[t]
 \centering
 \includegraphics[width=0.78\linewidth]{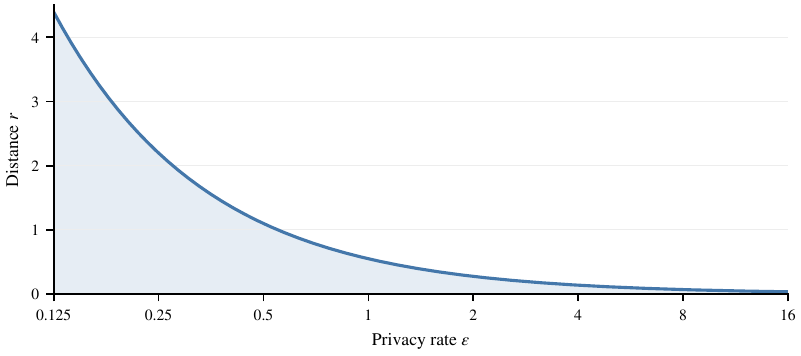}
 \caption{The shaded region satisfies $r \leq \log 3/(2\eps)$,
 guaranteeing at most $75\%$ identification accuracy on $\eps D$-private matching.}
 \label{fig:privacy-swap-calibration}
\end{figure}

Write $r=\|\vx-\vx'\|_2$. Both points change under the exchange, so
$D(X_0,X_1)=2r$. By definition of metric privacy \eqref{eq:dataset-privacy}, for every matching
$\sigma$ with positive probability,
\[
 e^{-2\eps r}
 \le \frac{\Pr[M_Y(X_0)=\sigma]}{\Pr[M_Y(X_1)=\sigma]}
 \le e^{2\eps r}.
\]
With equal prior probabilities, this ratio is also the posterior odds
after observing $\sigma$. Consequently, even the optimal attacker has
\begin{equation}
\label{eq:swap-inference-bound}
 \Pr[\text{correct guess}]
 \le \frac{e^{2\eps r}}{1+e^{2\eps r}}.
\end{equation}
To limit identification accuracy to at most $75\%$, compared with $50\%$
before the release, it therefore suffices to choose
\begin{equation}
\label{eq:swap-calibration}
 r\le\frac{\log 3}{2\eps},
 \qquad\text{equivalently}\qquad
 \eps\le\frac{\log 3}{2r},
\end{equation}
as illustrated in Figure~\ref{fig:privacy-swap-calibration}.

\paragraph{Inference attack on Sonar.}
The hypothesis-testing experiment pools $111$ metal-cylinder and $97$ rock
observations, giving $N=208$ data points with $d=60$ coordinates in
$[0,1]$. We use all observations in the source dataset, without further
standardization.
The minimum, median, and maximum Euclidean distances between distinct
points are $0.1746$, $1.7786$, and $3.5295$, respectively. These describe
the observed dataset; possible data points in the underlying population
can also lie between the observed points.

At $\eps=16$, $\eps D$-private matching guarantees accuracy at most
$75\%$ for exchanges between points separated by at most $r = 0.0343$.
This is below the minimum observed distance, so $\eps=16$ is large under this attack model. The same guarantee at the median observed
distance requires $\eps\le0.309$. This weakness should be viewed in light of the attacker's
extensive knowledge; both candidate points and all other individuals'
data, as well as the privacy mechanism itself, are already known, leaving only the two-way identity ambiguity.
In applications where an
attacker has less side information or observes only the test statistic, this
inference would be substantially more challenging. Thus, $\eps=16$ can remain a
useful choice under less informed adversaries, limiting their ability to distinguish between similar
data points while preserving evidence for the two-sample test.

\section{Proofs of Results}
\label{app:matching-perturbation}

\subsection{Matching Stability}

\begin{lemma}[Matching stability]
\label{lem:matching-stability}
Let $\mC\in\R^{n\times n}$ and let $\widehat{\mC}=\mC+\mE$.
Suppose $\widehat\sigma$ minimizes the matching cost for $\widehat{\mC}$.
Writing $u(\sigma)=\sum_i\emC_{i,\sigma(i)}$ and
$\OPT=\min_{\sigma\in S_n}u(\sigma)$, we have
\begin{equation}\label{eq:matching-stability}
 0\le\err(\widehat\sigma)
 \le\sum_{i=1}^n\bigl(\max_j\emE_{ij}-\min_j\emE_{ij}\bigr).
\end{equation}
The analogous bound with the sum of column ranges also holds.
\end{lemma}

\begin{proof}
Let $\sigma^\star$ minimize the true cost. Optimality for
$\widehat{\mC}$ gives
\[
 u(\widehat\sigma)+\sum_i\emE_{i,\widehat\sigma(i)}
 \le u(\sigma^\star)+\sum_i\emE_{i,\sigma^\star(i)}.
\]
Consequently,
\[
 \err(\widehat\sigma)
 \le\sum_i\bigl(\emE_{i,\sigma^\star(i)}
                 -\emE_{i,\widehat\sigma(i)}\bigr)
 \le\sum_i\bigl(\max_j\emE_{ij}-\min_j\emE_{ij}\bigr).
\]
Both permutations also select exactly one entry from each column, so the
same difference is bounded by the sum of the column ranges. Nonnegativity
follows from the definition of $\OPT$.
\end{proof}

In particular, a row or column offset adds the same constant to every
feasible matching cost and leaves all optimizers and excess costs unchanged.

\subsection{Original Points}

\begin{proof}[Proof of Theorem~\ref{thm:original-points}]
The Laplace mechanism gives local $\eps$-metric privacy for each report.
Parallel composition and post-processing give $\eps D$-privacy for the
output matching.

For utility, the cost error is
\[
 \emE_{ij}=\|\vx_i+\mathbf{Z}_i-\vy_j\|_2-\|\vx_i-\vy_j\|_2.
\]
The reverse triangle inequality gives
$|\emE_{ij}|\le\|\mathbf{Z}_i\|_2$, hence the error range in row $i$
is at most $2\|\mathbf{Z}_i\|_2$. Lemma~\ref{lem:matching-stability}
therefore implies, for every noise realization,
\[
 \err(\widehat\sigma)\le2\sum_{i=1}^n\|\mathbf{Z}_i\|_2.
\]
Since $\E\|\mathbf{Z}_i\|_2=d/\eps$, taking expectations gives
\[
 \E[\err(\widehat\sigma)]
 \le2\sum_{i=1}^n\E\|\mathbf{Z}_i\|_2=\frac{2nd}{\eps}. \qedhere
\]
\end{proof}

\subsection{Distance Vectors}

\begin{proof}[Proof of Theorem~\ref{thm:distance-vectors}]
For every public candidate $\vy_j$, the reverse triangle inequality gives
\[
 \bigl|\|\vx-\vy_j\|_2-\|\vx'-\vy_j\|_2\bigr|
 \le\|\vx-\vx'\|_2.
\]
Thus $\Phi_Y$ is $n^{1/p}$-Lipschitz into $\ell_p$:
\[
 \|\Phi_Y(\vx)-\Phi_Y(\vx')\|_p
 \le n^{1/p}\|\vx-\vx'\|_2.
\]
The chosen noise therefore gives local $\eps$-metric privacy. Parallel
composition and post-processing give $\eps D$-privacy for the output matching.

Write $Z_{ij}=[\mathbf{Z}_i]_j$. The cost error in row $i$ is precisely
$\mathbf{Z}_i$, so Lemma~\ref{lem:matching-stability} yields
\[
 \err(\widehat\sigma)
 \le\sum_{i=1}^n\bigl(\max_j Z_{ij}-\min_j Z_{ij}\bigr)
 \le2\sum_{i=1}^n\|\mathbf{Z}_i\|_\infty
 \le2\sum_{i=1}^n\|\mathbf{Z}_i\|_p.
\]
The $\ell_p$ noise magnitude satisfies
\[
 \|\mathbf{Z}_i\|_p\sim\operatorname{Gamma}
     \!\left(n,\text{rate }\frac{\eps}{n^{1/p}}\right),
 \qquad \E\|\mathbf{Z}_i\|_p=\frac{n^{1+1/p}}{\eps}.
\]
Since the noise vectors are identically distributed, taking expectations gives
\[
 \E[\err(\widehat\sigma)]
 \le2n\,\E\|\mathbf{Z}_1\|_p
 =\frac{2n^{2+1/p}}{\eps}.
\]

Since the upper
bound is minimized at $p=\infty$,
$\E\|\mathbf{Z}_1\|_\infty=n/\eps$, giving $2n^2/\eps$.
\end{proof}

\subsection{Projected Representation}
\label{app:projected-points-proof}

\begin{proof}[Proof of Theorem~\ref{thm:projected-points}]
When $s=d$, this is the original-point mechanism. For $s<d$, orthogonality
and the reverse triangle inequality give
\[
\begin{aligned}
 \|f_Y(\vx)-f_Y(\vx')\|_2^2
 &=\|P_W(\vx-\vx')\|_2^2
   +\bigl(\|P_{W^\perp}\vx\|_2-\|P_{W^\perp}\vx'\|_2\bigr)^2\\
 &\le\|P_W(\vx-\vx')\|_2^2
      +\|P_{W^\perp}(\vx-\vx')\|_2^2
 =\|\vx-\vx'\|_2^2.
\end{aligned}
\]
Adding Laplace noise therefore satisfies local $\eps$-metric privacy, and
composition and post-processing give $\eps D$-privacy for the matching.

Distance preservation and the reverse triangle inequality give
\[
 |\emE_{ij}|=|\widehat{\emC}_{ij}-\emC_{ij}|
 \le\|\mathbf{Z}_i\|_2.
\]
Applying Lemma~\ref{lem:matching-stability} and taking expectations gives
\[
 \E[\err(\widehat\sigma)]
 \le2\sum_{i=1}^n\E\|\mathbf{Z}_i\|_2
 =\frac{2ns}{\eps}.
\]
Finally, $\vy_1=0$ gives $r\le n-1$, so
$s=\min\{r+1,d\}\le\min\{n,d\}$.
\end{proof}

\subsection{Gumbel Matching}
\label{app:gumbel-matching-proof}

We prove the bound for $n\times n$ matching under a condition on the
cost changes within a row. This covers both Euclidean and squared
Euclidean costs.

\begin{theorem}
\label{thm:gumbel-row-bound}
Suppose each cost has the form $\emC_{ij}=\emC_{ij}(\vx_i)$ and, for
some $L>0$, satisfies
\begin{equation}\label{eq:gumbel-row-condition}
 \left|\bigl(\emC_{ij}(\vx)-\emC_{ij}(\vx')\bigr)
       -\bigl(\emC_{ik}(\vx)-\emC_{ik}(\vx')\bigr)\right|
 \le L\|\vx-\vx'\|_2
\end{equation}
for all $i,j,k\in[n]$ and $\vx,\vx'\in\R^d$.
For independent $G_{ij}\sim\operatorname{Gumbel}(0,1)$, the matching
\[
 \widehat\sigma\in\argmax_{\sigma\in S_n}
    \sum_{i=1}^n\left(-\emC_{i,\sigma(i)}
                         +\frac{L}{\eps}G_{i,\sigma(i)}\right)
\]
satisfies $\eps D$-privacy and
\[
 \E[\err(\widehat\sigma)]\le\frac{nL\log n}{\eps}.
\]
\end{theorem}

\begin{proof}
Set $\gamma=L/\eps$. First suppose $X,X'$ differ only in row $i$.
Condition on all noise outside that row, denoted $G_{-i}$. For each
$j\in[n]$, let
\[
 B_i(j)=\max_{\sigma\in S_n:\,\sigma(i)=j}
       \sum_{k\ne i}\bigl(-\emC_{k,\sigma(k)}
                              +\gamma G_{k,\sigma(k)}\bigr),
\]
and let $\sigma^{(j)}$ denote the corresponding maximizing permutation, using a fixed rule for any tie-breaking. Both $B_i(j)$ and $\sigma^{(j)}$ depend only on rows not equal to $i$ and $G_{-i}$, so they are the same for $X$ and $X'$.

Let $J = \widehat \sigma(i)$. Then $\widehat \sigma = \sigma^{(J)}$, and
\[
 J=\argmax_{j\in[n]}
       \bigl(-\emC_{ij}(\vx_i)+B_i(j)+\gamma G_{ij}\bigr).
\]
Define 
\[
 p_j(\vx_i) = \Pr[J = j \mid G_{-i}].
\]
The Gumbel-max trick \citep{mena2018learning} gives the conditional probabilities
\[
 p_j(\vx_i)=
 \frac{\exp\!\bigl((-\emC_{ij}(\vx_i)+B_i(j))/\gamma\bigr)}
 {\sum_{k=1}^n
    \exp\!\bigl((-\emC_{ik}(\vx_i)+B_i(k))/\gamma\bigr)}.
\]
Set $\delta_j=\emC_{ij}(\vx_i')-\emC_{ij}(\vx_i)$.
Condition~\eqref{eq:gumbel-row-condition} bounds their range by
$L\|\vx_i-\vx_i'\|_2$. Therefore
\[
 \frac{p_j(\vx_i)}{p_j(\vx_i')}
 =\frac{e^{\delta_j/\gamma}}
        {\sum_{k=1}^n p_k(\vx_i')e^{\delta_k/\gamma}}
 \le\exp\!\left(\frac{\max_k\delta_k-\min_k\delta_k}{\gamma}\right)
 \le e^{\eps\|\vx_i-\vx_i'\|_2}.
\]
For any event $A\subseteq S_n$, the output $\sigma^{(J)}$ belongs to $A$, which means
\[
\begin{aligned}
\Pr[M_Y(X)\in A\mid G_{-i}]
&=\sum_{j:\,\sigma^{(j)}\in A}p_j(\vx_i)\\
&\le
e^{\eps\|\vx_i-\vx_i'\|_2}
\sum_{j:\,\sigma^{(j)}\in A}p_j(\vx_i')\\
&=
e^{\eps\|\vx_i-\vx_i'\|_2}
\Pr[M_Y(X')\in A\mid G_{-i}].
\end{aligned}
\]
Since $G_{-i}$ is independent of $X,X'$, integrating over it on both sides gives
\[
\Pr[M_Y(X)\in A]
\le
e^{\eps\|\vx_i-\vx_i'\|_2}
\Pr[M_Y(X')\in A].
\]
For arbitrary $X$ and $X'$, replacing their records one at a time yields
\[
\Pr[M_Y(X)\in A]
\le
\exp\!\left(\eps\sum_{i=1}^n\|\vx_i-\vx_i'\|_2\right)
\Pr[M_Y(X')\in A]
=
e^{\eps D(X,X')}
\Pr[M_Y(X')\in A],
\]
establishing $\eps D$-privacy.

For utility, fix a minimum-cost matching $\sigma^\star$ independently of the
noise. Optimality for the perturbed scores gives
\[
 \err(\widehat\sigma)
 \le\gamma\sum_{i=1}^n
       \bigl(G_{i,\widehat\sigma(i)}-G_{i,\sigma^\star(i)}\bigr).
\]
The standard Gumbel cumulative distribution function shows that
$\max_{j\in[n]}G_{ij}$ has the same distribution as $G_{i1}+\log n$.
Consequently,
\[
 \E[\err(\widehat\sigma)]
 \le\gamma\sum_{i=1}^n
       \left(\E\max_{j\in[n]}G_{ij}-\E G_{i,\sigma^\star(i)}\right)
 =\gamma n\log n=\frac{L n\log n}{\eps}. \qedhere
\]
\end{proof}

\begin{proof}[Proof of Theorem~\ref{thm:gumbel-matching}]
For Euclidean costs, the reverse triangle inequality gives
\[
 \bigl|\|\vx-\vy_j\|_2-\|\vx'-\vy_j\|_2\bigr|
 \le\|\vx-\vx'\|_2.
\]
Thus \eqref{eq:gumbel-row-condition} holds with $L=2$, and
Theorem~\ref{thm:gumbel-row-bound} proves the claims.
\end{proof}

\subsection{Privacy Lower Bound}
\label{app:privacy-lower-bound-proof}

\begin{proof}[Proof of Theorem~\ref{thm:privacy-lower-bound}]
Choose points $\vv_1,\ldots,\vv_k$ in the Euclidean unit ball with mutual
distances greater than $1/2$, where
\[
 k=\min\{n,2^d\},\qquad
 h=\frac{\log k}{6\eps},\qquad
 G=\floor{n/k},\qquad N=Gk.
\]
Such points exist: a maximal set with separation greater than $1/2$ has
radius-$1/2$ balls covering the unit ball, so volume comparison gives at
least $2^d$ points. Since $k\ge2$ and $n-N<k\le N$, we have $N\ge n/2$.

Partition the first $N$ indices into $G$ groups $B_1,\ldots,B_G$ of size
$k$. Let $\ve_1$ denote the first coordinate vector and set
$\vc_g=10hg\ve_1$. For $a=1,\ldots,k$, place the public point with index
$j=(g-1)k+a$ at
\[
 \vy_j=\vc_g+h\vv_a+\vu_j,\qquad \|\vu_j\|_2\le h/16.
\]
Place each remaining public point at a separate location
$10h(G+a)\ve_1+\vu_{N+a}$, for $a=1,\ldots,n-N$, with the same bound on
its perturbation. Every full group has diameter at most $17h/8$.
Distinct points within a group are separated by more than
$h/2-2h/16=3h/8$. Points in different groups, including the remaining
single points, are separated by at least
$10h-2(17h/16)=63h/8$. Thus all distinct public points are separated by
at least $3h/8$.

Fix offsets $\vb_1,\ldots,\vb_n$ with $\|\vb_i\|_2\le h/16$.
Let $\Pi$ be a uniformly random permutation within each full group,
independently across groups, fixing the remaining indices. For each possible
permutation $\pi$, define a private input by
\[
 \vx_i^\pi=\vy_{\pi(i)}+\vb_i,\qquad i=1,\ldots,n.
\]
All public perturbations and private offsets stay fixed across this family.
The cost of the planted partner $\pi(i)$ is $\|\vb_i\|_2\le h/16$,
whereas every other partner costs at least $3h/8-h/16=5h/16$.
Consequently $\pi$ is the unique optimal matching, and every matching
$\sigma$ on this input satisfies
\begin{equation}\label{eq:lower-bound-cost-margin}
 \err(\sigma)
 \ge\frac h4\sum_{i=1}^n\1_{\{\sigma(i)\ne\pi(i)\}}.
\end{equation}

Fix any $\eps D$-private matching mechanism $M_Y$, and write
$\widehat\sigma=M_Y(X^\Pi)$. Consider a row $i$ in a full group $B_g$.
For each $a\in B_g$, write $\Pr_a$ for probability conditional on
$\Pi(i)=a$, averaging over the other permutation choices and the mechanism.
For distinct $a,b\in B_g$, exchanging the destination labels $a,b$ is a
bijection between the permutations with $\pi(i)=a$ and those with
$\pi(i)=b$. The paired private inputs differ in exactly two rows,
and their fixed offsets cancel, giving
\[
 D(X^\pi,X^{\pi'})=2\|\vy_a-\vy_b\|_2\le17h/4.
\]
Averaging the metric privacy inequality over these paired inputs gives,
for every event $A$ concerning $\widehat\sigma(i)$,
\[
 \Pr_a(A)\le\beta\Pr_b(A),\qquad
 \beta=\exp(17\eps h/4)\le k.
\]
In particular, for each $a\in B_g$,
\[
 \sum_{b\in B_g}\Pr_b\{\widehat\sigma(i)=a\}
 \ge\left(1+\frac{k-1}{\beta}\right)
       \Pr_a\{\widehat\sigma(i)=a\}.
\]
Summing over $a\in B_g$ and using disjointness of the possible assigned
labels yields
\[
 \left(1+\frac{k-1}{\beta}\right)
 \sum_{a\in B_g}\Pr_a\{\widehat\sigma(i)=a\}
 \le\sum_{b\in B_g}\sum_{a\in B_g}
          \Pr_b\{\widehat\sigma(i)=a\}\le k.
\]
Since $\Pi(i)$ is uniform on $B_g$, it follows that
\[
 \Pr\{\widehat\sigma(i)\ne\Pi(i)\}
 \ge\frac{k-1}{\beta+k-1}
 \ge\frac{k-1}{2k-1}\ge\frac13.
\]
Assignments outside the group also count as incorrect, so this argument
places no restriction on the mechanism's choice of partners.

Taking expectations in \eqref{eq:lower-bound-cost-margin} and summing over
the $N$ grouped rows gives
\[
 \E_{\Pi,M_Y}[\err(\widehat\sigma)]
 \ge\frac{hN}{12}
 \ge\frac{nh}{24}
 =\frac{n\log k}{144\eps}.
\]
Finally,
\[
 \log k=\min\{\log n,d\log2\}
 \ge(\log2)\min\{\log n,d\},\qquad
 \frac{\log2}{144}>\frac1{256}.
\]
At least one private input in the finite family therefore has expected
excess cost at least the claimed bound. The public construction was fixed
before choosing the mechanism.

The argument holds for every allowed public perturbation and private offset.
In particular, they can vary over open balls without changing the bound;
the construction does not require coincident private and public points or
exact distance symmetries.
\end{proof}

\section{Generalizations to Other Settings}
\label{app:mechanism-generalizations}

\subsection{Upper Bounds for Squared Euclidean Cost}
\label{app:squared-matching}

The squared objective allows us to discard terms that are constant within
a cost row. This simplifies the projected representation and lets us
calibrate distance-vector noise without bounding $X$. Let
\[
 \vc=\frac1n\sum_j\vy_j,\qquad
 W=\operatorname{span}\{\vy_j-\vc:j\in[n]\},\qquad r=\dim W.
\]
We use $\operatorname{diam}(Y)=\max_{j,k}\|\vy_j-\vy_k\|_2$.

\paragraph{Original points.}
Release $\widetilde\vx_i=\vx_i+\mathbf Z_i$ with the same
$d$-dimensional radial Laplace noise as in
Theorem~\ref{thm:original-points}, and minimize
$\sum_i\|\widetilde\vx_i-\vy_{\sigma(i)}\|_2^2$.

\paragraph{Projected representation.}
Let $\mB\in\R^{d\times r}$ have orthonormal columns spanning $W$,
and write $\vu_i=\mB^\top(\vx_i-\vc)$ and
$\vv_j=\mB^\top(\vy_j-\vc)$. Squared costs decompose as
\[
 \|\vx_i-\vy_j\|_2^2
 =\|\vu_i-\vv_j\|_2^2+
     \|P_{W^\perp}(\vx_i-\vc)\|_2^2.
\]
The last term is a row constant and does not affect the optimal
assignment. Projected representation therefore releases
$\widetilde\vu_i=\vu_i+\mathbf Z_i$ with radial Laplace noise in
$\R^r$, and minimizes $\sum_i\|\widetilde\vu_i-\vv_{\sigma(i)}\|_2^2$.

\paragraph{Distance vectors.}
Use the centered squared-cost representation
\begin{equation}\label{eq:squared-cost-representation}
 [h_Y(\vx)]_j=\|\vx-\vy_j\|_2^2-\|\vx-\vc\|_2^2
       =\|\vy_j-\vc\|_2^2-2\langle\vx-\vc,\vy_j-\vc\rangle.
\end{equation}
Subtracting the same term from every entry in a row leaves the matching
unchanged. Each record releases $h_Y(\vx_i)+\mathbf Z_i$, with density
\[
 p_{\mathbf Z_i}(\vz)\propto
       \exp\!\left(-\frac{\eps\|\vz\|_\infty}
                         {2\max_j\|\vy_j-\vc\|_2}\right).
\]
Centering gives a noise scale determined by $\max_j\|\vy_j-\vc\|_2$,
which can be smaller than $\operatorname{diam}(Y)$.

\paragraph{Gumbel.}
Draw independent standard Gumbels and return
\begin{equation}\label{eq:squared-gumbel}
 \widehat\sigma\in\argmax_{\sigma\in S_n}
   \sum_i\left(-\|\vx_i-\vy_{\sigma(i)}\|_2^2
                   +\frac{2\operatorname{diam}(Y)}{\eps}G_{i,\sigma(i)}\right).
\end{equation}
Theorem~\ref{thm:gumbel-row-bound} applies with
$L=2\operatorname{diam}(Y)$, since
\[
\begin{aligned}
 &\left|\bigl(\|\vx-\vy_j\|_2^2-\|\vx'-\vy_j\|_2^2\bigr)
       -\bigl(\|\vx-\vy_k\|_2^2-\|\vx'-\vy_k\|_2^2\bigr)\right|\\
 &\qquad
 =2|\langle\vx-\vx',\vy_j-\vy_k\rangle|
 \le2\operatorname{diam}(Y)\|\vx-\vx'\|_2.
\end{aligned}
\]

\begin{proposition}
\label{prop:squared-upper-bounds}
For $n\ge2$ and $\operatorname{diam}(Y)>0$, the four mechanisms satisfy
$\eps D$-privacy and the following expected excess-cost upper bounds
under squared Euclidean cost, where
$a_n(s)=\min\{s,2\sqrt{s\log n}\}$:
\begin{center}
\begin{tabular}{ll}
 \toprule
 Mechanism & Upper bound on $\E[\err(\widehat\sigma)]$\\
 \midrule
 Original points & $2n\operatorname{diam}(Y)a_n(d)/\eps$\\
 Projected representation & $2n\operatorname{diam}(Y)a_n(r)/\eps$\\
 Distance vectors & $4n^2\max_j\|\vy_j-\vc\|_2/\eps$\\
 Gumbel & $2n\operatorname{diam}(Y)\log n/\eps$\\
 \bottomrule
\end{tabular}
\end{center}
\end{proposition}

\begin{proof}
Projection is 1-Lipschitz, and $h_Y$ is
$2\max_j\|\vy_j-\vc\|_2$-Lipschitz in the infinity norm, giving privacy
for their Laplace reports. For Original points, expanding the perturbed
squared costs leaves the error $-2\langle\mathbf Z_i,\vy_j\rangle$
after removing a row constant. Lemma~\ref{lem:matching-stability} bounds
excess cost by the sum of the row oscillations
\[
 2\max_{j,k}\langle\mathbf Z_i,\vy_j-\vy_k\rangle.
\]
Recall that radial Laplace noise in $\R^s$ has the form
$\mathbf Z=R\bm\Theta$, with $\E R=s/\eps$ and an independent
uniform direction $\bm\Theta\in\sS^{s-1}$.
For a unit direction, the maximum inner product with a candidate
difference is at most $\operatorname{diam}(Y)$. The sphere's moment
formula also gives
\[
 \E\exp(t\langle\bm\Theta,\vv\rangle)
 \le\exp\!\left(\frac{t^2\|\vv\|_2^2}{2s}\right).
\]
Applying this to the at most $n^2$ differences of norm at most
$\operatorname{diam}(Y)$, then optimizing the exponential bound, yields
\[
 \E\max_{j,k}\langle\bm\Theta,\vy_j-\vy_k\rangle
 \le\operatorname{diam}(Y)
       \min\left\{1,2\sqrt{\frac{\log n}{s}}\right\}.
\]
Multiplying by $2\E R$ and summing over rows gives the Original points
bound with $s=d$. The same argument with $s=r$ and $\vv_j$ in place
of $\vy_j$ gives the Projected representation bound, since projection
preserves distances between points in $Y$. For Distance vectors, the
expected infinity norm of each noise row is
$2n\max_j\|\vy_j-\vc\|_2/\eps$; its oscillation is at most twice
that norm. Summing the stability bound gives
$4n^2\max_j\|\vy_j-\vc\|_2/\eps$. The Gumbel guarantee follows from
Theorem~\ref{thm:gumbel-row-bound} with $L=2\operatorname{diam}(Y)$.
\end{proof}

\subsection{Matching When Both Datasets Are Private}
\label{app:both-private-matching}

Suppose both datasets are private, while their sizes and indices remain
public. Extend the privacy metric to
\begin{equation}
\label{eq:both-private-metric}
 D_{\mathrm{both}}\bigl((X,Y),(X',Y')\bigr)
 =D(X,X')+D(Y,Y').
\end{equation}
We require the released matching $M(X,Y)$ to satisfy
\[
 \Pr[M(X,Y)\in A]
 \le e^{\eps D_{\mathrm{both}}((X,Y),(X',Y'))}
       \Pr[M(X',Y')\in A].
\]
The constructions below give the expected excess-cost bounds in
Table~\ref{tab:both-private-bounds}.

\begin{table}[htbp]
 \centering\small
 \caption{Upper bounds on $\E[\err(\widehat\sigma)]$ when both datasets
 are private under \eqref{eq:both-private-metric}. The dash indicates
 that the projection construction has no direct extension.}
 \label{tab:both-private-bounds}
 \begin{tabular}{ll}
  \toprule
  Mechanism & Euclidean cost\\
  \midrule
  Original points & $4nd/\eps$\\
  Projected representation & ---\\
  Distance vectors & $4n^2/\eps$\\
  Gumbel & $2n\log n/\eps$\\
  \bottomrule
 \end{tabular}
\end{table}

\paragraph{Original points.}
Independently perturb every point in both datasets by $d$-dimensional
radial Laplace noise with rate $\eps$, and compute a matching from the
reports. Their joint release satisfies $\eps D_{\mathrm{both}}$-privacy.
For Euclidean costs, the reverse triangle inequality gives
\[
 \err(\widehat\sigma)
 \le 2\sum_i\|\widetilde\vx_i-\vx_i\|_2
      +2\sum_j\|\widetilde\vy_j-\vy_j\|_2,
\]
whose expectation is at most $4nd/\eps$.

\paragraph{Projected representation.}
The representation in \eqref{eq:projected-representation} depends on
the exact affine span of $Y$. When $Y$ is private, that span also
requires protection, so the existing mechanism and its guarantee do
not apply by default.

\paragraph{Distance vectors.}
A central mechanism can protect both datasets by adding independent
noise to the rows and columns of the cost matrix
$\mC$ in \eqref{eq:matching-cost}.
Draw mutually independent noise vectors for the rows of
$\mZ^{\mathrm{row}}$ and the columns of $\mZ^{\mathrm{col}}$,
each with density
\[
 p(\vz)\propto\exp(-\eps\|\vz\|_\infty),
 \qquad \vz\in\R^n.
\]
Return a minimum-cost matching for
$\mC+\mZ^{\mathrm{row}}+\mZ^{\mathrm{col}}$.
A change to $\vx_i$ affects only row $i$, whose infinity-norm change
is at most $\|\vx_i-\vx_i'\|_2$. Conditioning on the column noise,
the row noise therefore gives the required privacy factor.
The same argument with rows and columns exchanged protects a change
to $\vy_j$. Replacing the points one at a time establishes
$\eps D_{\mathrm{both}}$-privacy. The stability argument in
Lemma~\ref{lem:matching-stability} bounds excess cost by the sum of
the row oscillations of $\mZ^{\mathrm{row}}$ and the column oscillations
of $\mZ^{\mathrm{col}}$. Each noise vector has expected infinity norm
$n/\eps$, giving $\E[\err(\widehat\sigma)]\le4n^2/\eps$.

\paragraph{Gumbel.}
For Euclidean costs, the mechanism needs no change. The proof of Theorem~\ref{thm:gumbel-row-bound}
protects changes to $X$. Transposing the cost and noise matrices gives
the same proof for changes to $Y$: condition on noise outside the
changed column and apply the Gumbel-max argument to its possible
partners. Replacing rows and then columns proves
$\eps D_{\mathrm{both}}$-privacy, while the utility bound remains
$2n\log n/\eps$.

\clearpage
\section{Details of the Numerical Experiments}
\label{app:exp-numerical-details}

\subsection{Datasets and Feature Treatment}
\label{app:exp-datasets}

Table~\ref{tab:exp-datasets} lists the sixteen datasets. The clustering
suite is fixed at version 1.2.0. For MNIST and Fashion-MNIST, we use its
10,000-image test subsets. Iris uses the suite's copy of the R dataset.
The first three geometries and Worms-64 are synthetic; the remaining
datasets contain real measurements or images. These sources provide
different dimensions, shapes, and feature distributions for matching,
independently of their original clustering or classification labels.

\begin{table}[t]
 \centering\small
 \caption{Dataset sizes before sampling, ambient dimensions, and original
 sources. All privacy sweeps use 64 points per side.}
 \label{tab:exp-datasets}
 \setlength{\tabcolsep}{5pt}
 \begin{tabular}{lrrl}
  \toprule
  Dataset & Available records & $d$ & Source\\
  \midrule
  Ring noisy & 1,050 & 2 & \citet{graves2010kernel}\\
  Circles & 4,000 & 2 & \citet{truszczynski2019circles}\\
  Chainlink & 1,000 & 3 & \citet{ultsch2005clustering}\\
  Iris & 150 & 4 & \citet{fisher1936measurements}\\
  Banknote & 1,372 & 4 & \citet{lohweg2012banknote}\\
  Ecoli & 336 & 7 & \citet{nakai1996ecoli}\\
  Yeast & 1,484 & 8 & \citet{nakai1991yeast}\\
  Glass & 214 & 9 & \citet{german1987glass}\\
  Wine & 178 & 13 & \citet{aeberhard1992wine}\\
  Statlog (image segmentation) & 2,310 & 19 & \citet{uci1990segmentation}\\
  WDBC & 569 & 30 & \citet{wolberg1993wdbc}\\
  Ionosphere & 351 & 34 & \citet{sigillito1989ionosphere}\\
  Sonar & 208 & 60 & \citet{sejnowski1988sonar}\\
  Worms-64 & 105,000 & 64 & \citet{sieranoja2019density}\\
  MNIST & 10,000 & 784 & \citet{lecun1998gradient}\\
  Fashion-MNIST & 10,000 & 784 & \citet{xiao2017fashion}\\
  \bottomrule
 \end{tabular}
\end{table}

For each parent collection, we fit the feature transform once, before
making public/private partitions. A varying numerical feature $z_j$ is
mapped to $(z_j-\mu_j)/\sigma_j$, using the empirical mean and
standard deviation of the pooled parent collection. A constant feature is centered to zero and retains
its coordinate. A binary feature with fixed source values $a_j,b_j$ is
mapped to $(z_j-a_j)/(b_j-a_j)$. Its weight is therefore independent of
the indicator's frequency in the sampled collection.

Wine, WDBC, Glass, Sonar, and Banknote use numerical standardization
for all features. Statlog uses it for varying numerical features and
retains its constant coordinates. Ecoli standardizes five numerical
scores and maps its two binary features to $\{0,1\}$. Yeast standardizes
seven numerical scores and maps its binary indicator to $\{0,1\}$.
Ionosphere retains the first binary coordinate and second zero
coordinate, and standardizes coordinates three through thirty-four.
Iris, the synthetic datasets, and both image datasets retain their
relative coordinate scales before the common RMS transformation.

As an example, a sample from Ecoli can be written schematically as
$(c_1,c_2,b_1,b_2,c_3,c_4,c_5)$, with numerical scores $c_j$ and binary
indicators $b_j$. After feature treatment, the corresponding vector is
\[
 \left(\frac{c_1-\mu_1}{\sigma_1},
       \frac{c_2-\mu_2}{\sigma_2},b_1,b_2,
       \frac{c_3-\mu_3}{\sigma_3},
       \frac{c_4-\mu_4}{\sigma_4},
       \frac{c_5-\mu_5}{\sigma_5}\right),
 \qquad b_1,b_2\in\{0,1\}.
\]
The common centering and RMS rescaling below then apply to the entire
vector, including the binary coordinates.

\subsection{Distance Scale, Projection Chains, and Nested Samples}
\label{app:exp-normalization}

For the $2n$ pooled, feature-treated vectors $\vv_1,\ldots,\vv_{2n}$ in an
actual instance, define
\begin{equation}
\label{eq:exp-rms-normalization}
 \bar \vv=\frac1{2n}\sum_{i=1}^{2n}\vv_i,\qquad
 a^2=\frac1{2n}\sum_{i=1}^{2n}\|\vv_i-\bar \vv\|_2^2,\qquad
 \vz_i=\frac{\vv_i-\bar \vv}{a}.
\end{equation}
The resulting pooled cloud has mean zero and RMS radius one. Feature
weights, center, and radius are treated as fixed parts of the offline
benchmark construction. The mechanism's privacy guarantee is evaluated
on these normalized inputs.

As discussed in Appendix~\ref{app:privacy-calibration}, the privacy comparison
at a one-record displacement $\rho$ is controlled by
$\exp(\eps\rho)$. In particular, a factor-$c$ guarantee extends to
normalized distance $\rho_c=\log(c)/\eps$. At $\eps=8$, the factor-two
distance is about 0.0866; at $\eps=16$, it is about 0.0433. Equation
\eqref{eq:exp-rms-normalization} expresses those distances in units of a
common RMS radius. For a fixed feature transform and radius $a$, a
feature-space displacement $\delta$ becomes $\delta/a$, giving privacy
factor $\exp(\eps\delta/a)$. Thus holding $\eps$ fixed after RMS
normalization holds the protection scale fixed relative to that radius.
This calibration motivates controlling the radius in both sweeps.

\paragraph{Fashion-MNIST projections.}
Each parent contains 1,600 distinct full-dimensional images. Write its
centered, RMS-normalized data matrix as $\mZ\in\R^{1600\times784}$.
For each of five independent chains, draw a Haar-distributed orthogonal
matrix $\mO\in\R^{784\times784}$. We obtain it by Gaussian QR
factorization with the signs fixed by a positive diagonal in the
triangular factor \citep{mezzadri2007random}. Let $\mO_d$ contain its first
$d$ columns. At each target dimension, set
\begin{equation}
\label{eq:exp-projection-normalization}
 \begin{aligned}
 \mV_d&=\mZ\mO_d,\qquad
 \bar \vv_d=\frac1{1600}\mV_d^\top\1,\\
 a_d^2&=\frac1{1600}\sum_i\|(\mV_d)_{i,:}^\top-\bar \vv_d\|_2^2,
 \qquad \mZ_d=\frac{\mV_d-\1\bar \vv_d^\top}{a_d}.
 \end{aligned}
\end{equation}
Here $(\mV_d)_{i,:}$ is row $i$ of the projected data matrix.
Every target retains the same 800 public and 800 private
records in each of five fixed partitions.

The prefixes $\mO_d$ form nested uniformly random subspaces. Their joint
subspace distribution agrees with successive uniform orthogonal
reductions from dimension 784, up to an orthogonal change of coordinates
within each retained space. All four matching mechanisms are invariant
in law under such coordinate rotations. For a fixed row vector $\vv$,
\[
 \E_{\mO}\|\vv\mO_d\|_2^2=\frac{d}{784}\|\vv\|_2^2.
\]
Multiplication by $\sqrt{784/d}$ would preserve squared length in
expectation. Dividing by the actual pooled RMS radius in
\eqref{eq:exp-projection-normalization} cancels that scalar and fixes
the radius of every realized projected cloud.

This also explains the equivalence to iterative rescaling. If a
reduction from $d$ to $d'$ is represented by an orthonormal matrix $\mA$
with $\mO_d\mA=\mO_{d'}$, the current centered data project to
$(\mZ\mO_{d'}-\1\bar \vv_{d'}^\top)/a_d$. Their new RMS radius
is $a_{d'}/a_d$, so the next normalization gives exactly $\mZ_{d'}$.
Intermediate scalar normalizations therefore cancel algebraically.
The experiment compares a family of projected matching geometries at
the same RMS radius, averaging five independent chains to account for
the choice of retained subspaces.

The target dimensions are $784,604,404,304,254$, every fifth dimension
from 204 down to 4, and every integer from 1 through 12: fifty-six
distinct values. Figure~\ref{fig:exp-dimension-and-size}(a) displays
the portion through 104. Table~\ref{tab:exp-fashion-values} includes
every measured dimension.

\paragraph{Banknote nested samples.}
Each parent contains 400 distinct records. We fit the four feature
means and standard deviations once on this parent. Each of five
partitions orders 200 public and 200 private records; for a target $n$,
we take the first $n$ records from each side. Thus smaller samples retain
their memberships when $n$ increases. For each selected pooled sample,
we recompute $\bar \vv_n$ and $a_n$ using
\eqref{eq:exp-rms-normalization}, keeping the feature weights fixed.
The same numerical $\eps=8$ then applies at RMS radius one throughout.
The changing sample size affects both the set of possible partners and
the realized geometry, with their overall distance scale controlled.

\subsection{Mechanism Calibration and Repeated Runs}
\label{app:exp-mechanisms}

For Euclidean costs, Original points adds $d$-dimensional radial
Laplace noise with density proportional to $\exp(-\eps\|\vz\|_2)$.
Projected representation adds the same type of noise in
$s=\min\{r+1,d\}$ coordinates, where $r$ is the affine dimension of
the public points. Distance vectors adds infinity-norm Laplace noise
with density proportional to $\exp(-\eps\|\vz\|_\infty)$ to each cost
row. Gumbel adds independent standard Gumbels with scale $2/\eps$ to
the negative edge costs and maximizes the perturbed assignment score.

For squared costs, Original points retains its $d$-dimensional report.
Projected representation uses the public affine span: for a public
orthonormal basis $\mB\in\R^{d\times r}$ and public center $\vc$,
it reports $\mB^\top(\vx_i-\vc)+\mathbf Z_i$ with $r$-dimensional radial
Laplace noise. The perpendicular squared norm is constant within a cost
row, so the assignment uses only these projected coordinates. Distance
vectors perturbs
\[
 h_j(\vx)=\|\vx-\vy_j\|_2^2-\|\vx-\vc\|_2^2
       =\|\vy_j-\vc\|_2^2-2\langle \vx-\vc,\vy_j-\vc\rangle,
 \qquad \vc=\frac1n\sum_j \vy_j,
\]
using density proportional to
$\exp[-\eps\|\vz\|_\infty/(2\max_j\|\vy_j-\vc\|_2)]$.
Gumbel's squared-cost noise scale is
$2\operatorname{diam}(Y)/\eps$. These calibrations give privacy in
the same recordwise Euclidean metric for both objectives; the
squared-cost bounds are given in Appendix~\ref{app:squared-matching}.

We determine the public affine dimension from the centered public
matrix, retaining singular values above $\max\{n,d\}\sigma_{\max}$
times machine precision, where $\sigma_{\max}$ is its largest singular value.

Whenever a projected report is the original-point report, we share the
entire realization, including its assignment. Such coincident curves
therefore represent the same randomized mechanism. For each fixed
geometry, unit-noise draws are shared across privacy budgets and
rescaled by $\eps$; mechanisms otherwise have independent random
channels. Each stated privacy guarantee concerns one execution at its
specified budget.

The twenty parent collections are independent draws from the finite
source dataset, and different parents may contain some of the same
records. Within a parent, all selected records are distinct and every
public/private partition is disjoint. Fifty runs estimate the mechanism
expectation conditional on that partition. We average those runs,
partitions, and projection chains before forming the twenty parent
means and their Student-$t$ intervals. Reported intervals are pointwise
and unadjusted for comparisons across settings.

\subsection{Numerical Values}
\label{app:exp-numerical-values}

The following tables give all sixteen datasets at three representative
budgets under each objective, and every measured dimension and sample
size in the two dedicated sweeps. Each mechanism entry reports a mean
and the half-width of its pointwise 95\% confidence interval. Total
excess costs are used except in the Banknote table, which reports
excess cost per pair.

\input{experiment_numerical_tables}

\clearpage
\subsection{Privacy-Budget Sweeps}
\label{app:exp-epsilon-sweeps}

The following figures give the full privacy-budget sweeps for all sixteen
benchmarks, with $n=64$ per side. Each panel uses its own vertical scale;
gray dotted lines give the random reference for the corresponding
dataset and objective. Appendix~\ref{app:exp-numerical-values} reports
means and confidence intervals at representative budgets.

\subsubsection{Euclidean Costs}
\label{app:exp-epsilon-euclidean}

\begin{figure}[h]
 \centering
 \includegraphics[width=\linewidth]{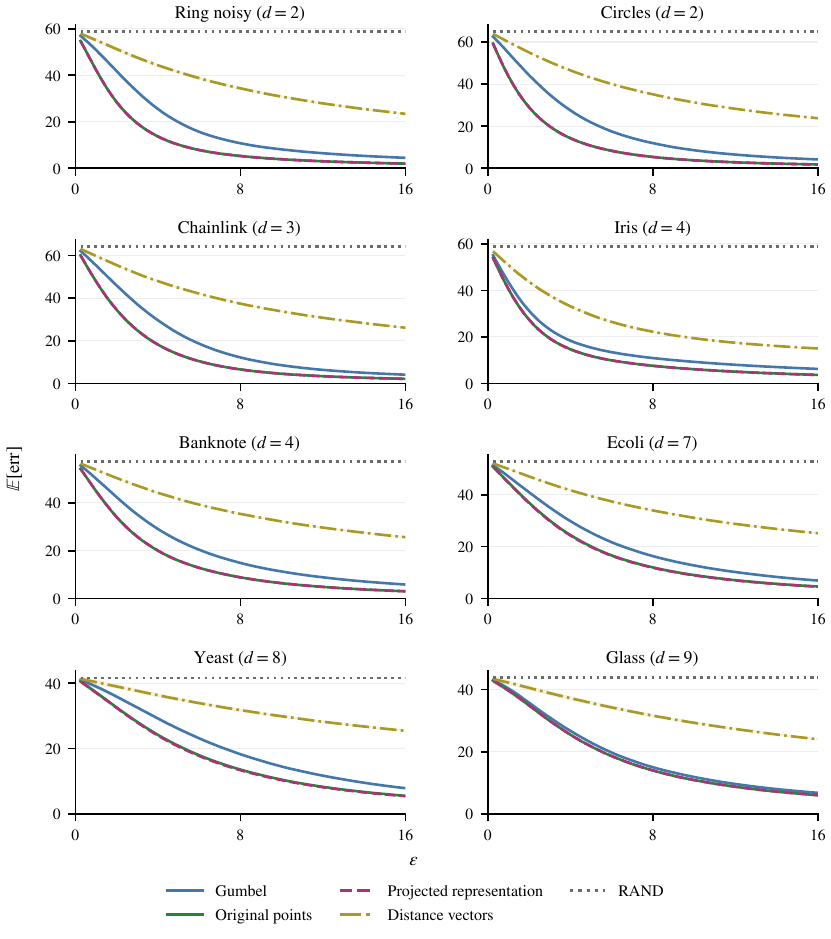}
 \caption{Euclidean excess matching cost across privacy budgets for the
 eight lowest-dimensional benchmarks.}
 \label{fig:exp-epsilon-euclidean-low}
\end{figure}

\begin{figure}[p]
 \centering
 \includegraphics[width=\linewidth]{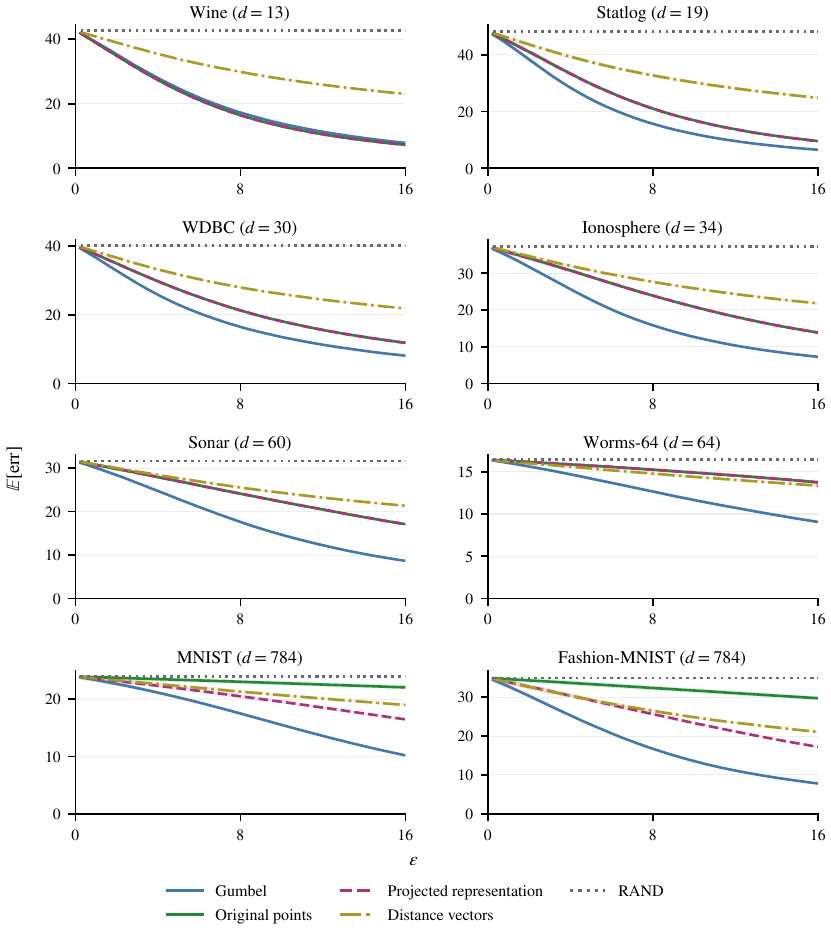}
 \caption{Euclidean excess matching cost across privacy budgets for the
 remaining eight benchmarks.}
 \label{fig:exp-epsilon-euclidean-high}
\end{figure}

\clearpage
\subsubsection{Squared Euclidean Costs}
\label{app:exp-epsilon-squared}

\begin{figure}[!htbp]
 \centering
 \includegraphics[width=\linewidth]{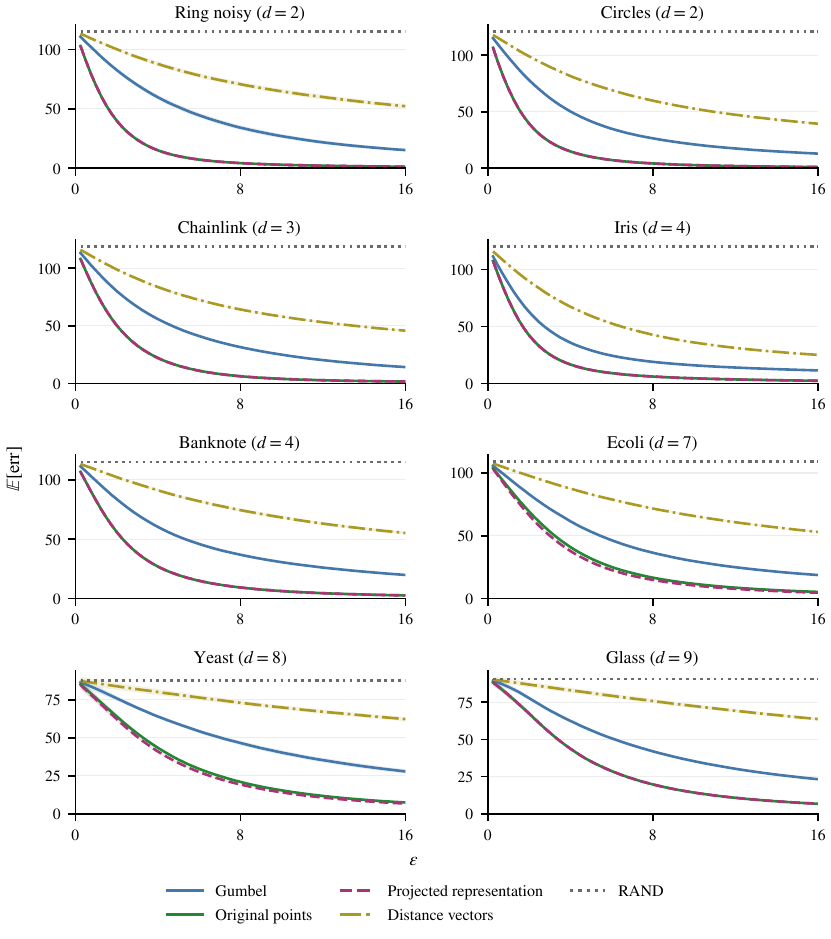}
 \caption{Squared Euclidean excess matching cost across privacy budgets
 for the eight lowest-dimensional benchmarks.}
 \label{fig:exp-epsilon-squared-low}
\end{figure}

\begin{figure}[p]
 \centering
 \includegraphics[width=\linewidth]{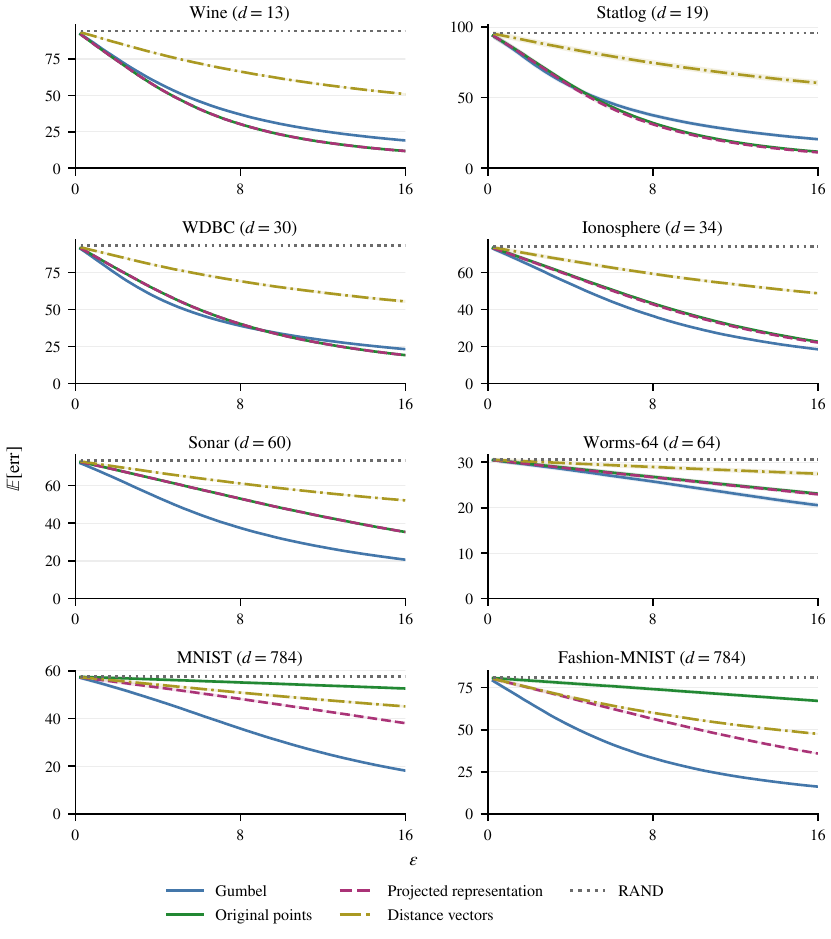}
 \caption{Squared Euclidean excess matching cost across privacy budgets
 for the remaining eight benchmarks.}
 \label{fig:exp-epsilon-squared-high}
\end{figure}

\clearpage
\section{Details of the Private Rank-Energy Experiment}
\label{app:exp-rank-details}

\subsection{Reference Grid and Null Distribution}

The grid consists of the unscrambled Halton points indexed 1 through 208,
using the first 60 primes as coordinate bases. It is independent of both
measurements and labels. Its affine rank is 47. Matching uses neither
cohort labels nor label-dependent constraints, and randomized tie-breaking
treats rows symmetrically in distribution.

Under $H_0$, the reference ranks are uniformly distributed over all
permutations of $Y$. We approximate the
resulting null distribution by Monte Carlo simulation. Since $T$ is
invariant under within-cohort permutations, each draw requires only a
uniformly chosen subset of $a$ reference points for the first cohort,
with the remaining $b$ assigned to the second.
With $J=9{,}999$ simulated statistics $T^{(1)},\ldots,T^{(J)}$, we calculate
\begin{equation}
\label{eq:exp-sonar-pvalue}
 \widehat p=\frac{1+\sum_{j=1}^J
              \1\{T^{(j)}\ge T(\widehat\sigma)\}}{J+1}.
\end{equation}
The observed private assignment stays fixed throughout this calculation.
Since every assignment uses all reference points once, the simulated
null statistics can be used for every method and run.

The figure's confidence intervals use the Student-$t$ distribution over
the 1,000 independent runs for each method and $\eps$, with the simulated
null sample fixed.

Runs are independent at each fixed $\eps$. The same underlying noise
draws are rescaled across $\eps$ to compare budgets; these comparisons
do not compose the privacy across independent runs. The privacy guarantee applies only to each
individual execution.

\end{document}

%% file: math_commands.tex
\usepackage{amsmath,amsfonts,bm}

\def\eqref#1{equation~\ref{#1}}

\def\floor#1{\lfloor #1 \rfloor}
\def\1{\bm{1}}

\def\eps{{\epsilon}}

\def\vb{{\bm{b}}}
\def\vc{{\bm{c}}}

\def\ve{{\bm{e}}}

\def\vu{{\bm{u}}}
\def\vv{{\bm{v}}}

\def\vx{{\bm{x}}}
\def\vy{{\bm{y}}}
\def\vz{{\bm{z}}}

\def\mA{{\bm{A}}}
\def\mB{{\bm{B}}}
\def\mC{{\bm{C}}}

\def\mE{{\bm{E}}}

\def\mO{{\bm{O}}}

\def\mV{{\bm{V}}}

\def\mZ{{\bm{Z}}}

\DeclareMathAlphabet{\mathsfit}{\encodingdefault}{\sfdefault}{m}{sl}
\SetMathAlphabet{\mathsfit}{bold}{\encodingdefault}{\sfdefault}{bx}{n}

\def\sR{{\mathbb{R}}}
\def\sS{{\mathbb{S}}}

\def\emC{{C}}

\def\emE{{E}}

\newcommand{\E}{\mathbb{E}}

\newcommand{\R}{\mathbb{R}}

\DeclareMathOperator*{\argmax}{arg\,max}
\DeclareMathOperator*{\argmin}{arg\,min}

%% file: experiment_numerical_tables.tex
\begin{table}[!htbp]
\centering\scriptsize
\setlength{\tabcolsep}{3pt}
\caption{Banknote size sweep, $d=4$, $\eps=8$. Mechanism entries are per-pair excess-cost means $\pm$ pointwise 95\% confidence half-widths. RAND is $(\mathrm{RAND}-\mathrm{OPT})/n$; $B/n=2\min\{4,\log n\}/8$.}
\label{tab:exp-banknote-values}
\begin{tabular}{rrrrrrr}
\toprule
$n$ & Gumbel & Original points & \shortstack{Projected\\representation} & Distance vectors & RAND & $B/n$\\
\midrule
20 & $0.162\,\pm\,0.003$ & $0.091\,\pm\,0.003$ & $0.091\,\pm\,0.003$ & $0.359\,\pm\,0.008$ & 0.696 & 0.749\\
40 & $0.205\,\pm\,0.003$ & $0.119\,\pm\,0.003$ & $0.119\,\pm\,0.003$ & $0.484\,\pm\,0.010$ & 0.822 & 0.922\\
60 & $0.232\,\pm\,0.002$ & $0.137\,\pm\,0.002$ & $0.137\,\pm\,0.002$ & $0.547\,\pm\,0.006$ & 0.883 & 1.000\\
80 & $0.252\,\pm\,0.002$ & $0.149\,\pm\,0.002$ & $0.149\,\pm\,0.002$ & $0.588\,\pm\,0.006$ & 0.924 & 1.000\\
100 & $0.265\,\pm\,0.002$ & $0.158\,\pm\,0.002$ & $0.158\,\pm\,0.002$ & $0.617\,\pm\,0.004$ & 0.953 & 1.000\\
120 & $0.277\,\pm\,0.003$ & $0.166\,\pm\,0.002$ & $0.166\,\pm\,0.002$ & $0.640\,\pm\,0.005$ & 0.976 & 1.000\\
140 & $0.285\,\pm\,0.002$ & $0.172\,\pm\,0.002$ & $0.172\,\pm\,0.002$ & $0.655\,\pm\,0.005$ & 0.992 & 1.000\\
160 & $0.294\,\pm\,0.002$ & $0.179\,\pm\,0.001$ & $0.179\,\pm\,0.001$ & $0.670\,\pm\,0.004$ & 1.007 & 1.000\\
180 & $0.301\,\pm\,0.002$ & $0.183\,\pm\,0.001$ & $0.183\,\pm\,0.001$ & $0.683\,\pm\,0.004$ & 1.019 & 1.000\\
200 & $0.307\,\pm\,0.002$ & $0.188\,\pm\,0.001$ & $0.188\,\pm\,0.001$ & $0.693\,\pm\,0.004$ & 1.030 & 1.000\\
\bottomrule
\end{tabular}
\end{table}

\begin{table}[p]
\centering\scriptsize
\setlength{\tabcolsep}{3pt}
\caption{Euclidean excess matching cost at $\eps=4$, $n=64$. Entries are means $\pm$ pointwise 95\% confidence half-widths over twenty parent collections. RAND denotes $\mathrm{RAND}-\mathrm{OPT}$.}
\label{tab:exp-bench-p1-eps4}
\begin{tabular}{lrrrrr}
\toprule
Dataset & Gumbel & Original points & \shortstack{Projected\\representation} & Distance vectors & RAND\\
\midrule
Ring noisy & $25.534\,\pm\,0.517$ & $13.661\,\pm\,0.337$ & $13.661\,\pm\,0.337$ & $44.320\,\pm\,0.752$ & 58.825\\
Circles & $27.534\,\pm\,0.178$ & $14.271\,\pm\,0.155$ & $14.271\,\pm\,0.155$ & $46.525\,\pm\,0.260$ & 65.009\\
Chainlink & $29.642\,\pm\,0.349$ & $18.220\,\pm\,0.314$ & $18.220\,\pm\,0.314$ & $47.962\,\pm\,0.500$ & 64.143\\
Iris & $18.372\,\pm\,0.347$ & $14.700\,\pm\,0.360$ & $14.700\,\pm\,0.360$ & $33.181\,\pm\,0.686$ & 58.935\\
Banknote & $29.061\,\pm\,0.450$ & $20.038\,\pm\,0.313$ & $20.038\,\pm\,0.313$ & $44.162\,\pm\,0.443$ & 57.155\\
Ecoli & $29.831\,\pm\,0.313$ & $24.482\,\pm\,0.275$ & $24.296\,\pm\,0.328$ & $41.697\,\pm\,0.367$ & 52.776\\
Yeast & $29.219\,\pm\,0.402$ & $24.193\,\pm\,0.281$ & $23.979\,\pm\,0.305$ & $36.331\,\pm\,0.615$ & 41.638\\
Glass & $26.717\,\pm\,0.334$ & $25.480\,\pm\,0.372$ & $25.480\,\pm\,0.372$ & $37.253\,\pm\,0.695$ & 43.824\\
Wine & $27.969\,\pm\,0.283$ & $27.355\,\pm\,0.298$ & $27.355\,\pm\,0.298$ & $35.426\,\pm\,0.297$ & 42.668\\
Statlog & $28.433\,\pm\,0.513$ & $33.398\,\pm\,0.622$ & $33.335\,\pm\,0.657$ & $39.350\,\pm\,0.731$ & 48.201\\
WDBC & $25.814\,\pm\,0.269$ & $29.746\,\pm\,0.319$ & $29.746\,\pm\,0.319$ & $33.190\,\pm\,0.309$ & 40.129\\
Ionosphere & $25.661\,\pm\,0.323$ & $30.743\,\pm\,0.490$ & $30.743\,\pm\,0.490$ & $32.017\,\pm\,0.464$ & 37.223\\
Sonar & $24.745\,\pm\,0.245$ & $27.940\,\pm\,0.371$ & $27.940\,\pm\,0.371$ & $28.412\,\pm\,0.350$ & 31.600\\
Worms-64 & $14.689\,\pm\,0.170$ & $15.853\,\pm\,0.176$ & $15.853\,\pm\,0.176$ & $15.572\,\pm\,0.182$ & 16.423\\
MNIST & $21.126\,\pm\,0.311$ & $23.480\,\pm\,0.394$ & $22.319\,\pm\,0.366$ & $22.619\,\pm\,0.361$ & 23.943\\
Fashion-MNIST & $25.312\,\pm\,0.360$ & $33.661\,\pm\,0.621$ & $30.369\,\pm\,0.562$ & $30.374\,\pm\,0.532$ & 34.937\\
\bottomrule
\end{tabular}
\end{table}

\begin{table}[p]
\centering\scriptsize
\setlength{\tabcolsep}{3pt}
\caption{Euclidean excess matching cost at $\eps=8$, $n=64$. Entries are means $\pm$ pointwise 95\% confidence half-widths over twenty parent collections. RAND denotes $\mathrm{RAND}-\mathrm{OPT}$.}
\label{tab:exp-bench-p1-eps8}
\begin{tabular}{lrrrrr}
\toprule
Dataset & Gumbel & Original points & \shortstack{Projected\\representation} & Distance vectors & RAND\\
\midrule
Ring noisy & $10.757\,\pm\,0.179$ & $5.273\,\pm\,0.103$ & $5.273\,\pm\,0.103$ & $34.399\,\pm\,0.654$ & 58.825\\
Circles & $11.942\,\pm\,0.101$ & $5.359\,\pm\,0.068$ & $5.359\,\pm\,0.068$ & $35.149\,\pm\,0.248$ & 65.009\\
Chainlink & $12.191\,\pm\,0.124$ & $6.568\,\pm\,0.097$ & $6.568\,\pm\,0.097$ & $37.409\,\pm\,0.441$ & 64.143\\
Iris & $10.927\,\pm\,0.120$ & $7.588\,\pm\,0.140$ & $7.588\,\pm\,0.140$ & $22.245\,\pm\,0.455$ & 58.935\\
Banknote & $14.909\,\pm\,0.274$ & $8.885\,\pm\,0.141$ & $8.885\,\pm\,0.141$ & $35.375\,\pm\,0.407$ & 57.155\\
Ecoli & $16.362\,\pm\,0.161$ & $12.025\,\pm\,0.156$ & $11.902\,\pm\,0.194$ & $33.942\,\pm\,0.349$ & 52.776\\
Yeast & $18.259\,\pm\,0.193$ & $13.565\,\pm\,0.201$ & $13.366\,\pm\,0.211$ & $31.768\,\pm\,0.486$ & 41.638\\
Glass & $15.002\,\pm\,0.189$ & $13.981\,\pm\,0.175$ & $13.981\,\pm\,0.175$ & $31.604\,\pm\,0.532$ & 43.824\\
Wine & $17.265\,\pm\,0.159$ & $16.434\,\pm\,0.158$ & $16.434\,\pm\,0.158$ & $29.867\,\pm\,0.295$ & 42.668\\
Statlog & $15.682\,\pm\,0.220$ & $21.052\,\pm\,0.398$ & $21.015\,\pm\,0.424$ & $32.750\,\pm\,0.662$ & 48.201\\
WDBC & $16.477\,\pm\,0.185$ & $21.256\,\pm\,0.222$ & $21.256\,\pm\,0.222$ & $27.949\,\pm\,0.259$ & 40.129\\
Ionosphere & $15.834\,\pm\,0.216$ & $23.912\,\pm\,0.308$ & $23.912\,\pm\,0.308$ & $27.659\,\pm\,0.402$ & 37.223\\
Sonar & $17.628\,\pm\,0.135$ & $24.114\,\pm\,0.305$ & $24.114\,\pm\,0.305$ & $25.539\,\pm\,0.289$ & 31.600\\
Worms-64 & $12.656\,\pm\,0.167$ & $15.231\,\pm\,0.179$ & $15.231\,\pm\,0.179$ & $14.774\,\pm\,0.181$ & 16.423\\
MNIST & $17.499\,\pm\,0.212$ & $23.018\,\pm\,0.379$ & $20.484\,\pm\,0.317$ & $21.307\,\pm\,0.327$ & 23.943\\
Fashion-MNIST & $16.724\,\pm\,0.190$ & $32.351\,\pm\,0.597$ & $25.625\,\pm\,0.432$ & $26.492\,\pm\,0.429$ & 34.937\\
\bottomrule
\end{tabular}
\end{table}

\begin{table}[p]
\centering\scriptsize
\setlength{\tabcolsep}{3pt}
\caption{Euclidean excess matching cost at $\eps=16$, $n=64$. Entries are means $\pm$ pointwise 95\% confidence half-widths over twenty parent collections. RAND denotes $\mathrm{RAND}-\mathrm{OPT}$.}
\label{tab:exp-bench-p1-eps16}
\begin{tabular}{lrrrrr}
\toprule
Dataset & Gumbel & Original points & \shortstack{Projected\\representation} & Distance vectors & RAND\\
\midrule
Ring noisy & $4.483\,\pm\,0.075$ & $1.994\,\pm\,0.047$ & $1.994\,\pm\,0.047$ & $23.424\,\pm\,0.479$ & 58.825\\
Circles & $4.182\,\pm\,0.055$ & $1.786\,\pm\,0.038$ & $1.786\,\pm\,0.038$ & $23.769\,\pm\,0.220$ & 65.009\\
Chainlink & $4.109\,\pm\,0.047$ & $2.161\,\pm\,0.028$ & $2.161\,\pm\,0.028$ & $26.132\,\pm\,0.411$ & 64.143\\
Iris & $6.242\,\pm\,0.047$ & $3.666\,\pm\,0.067$ & $3.666\,\pm\,0.067$ & $15.044\,\pm\,0.263$ & 58.935\\
Banknote & $5.886\,\pm\,0.078$ & $3.081\,\pm\,0.051$ & $3.081\,\pm\,0.051$ & $25.668\,\pm\,0.400$ & 57.155\\
Ecoli & $6.968\,\pm\,0.078$ & $4.658\,\pm\,0.054$ & $4.598\,\pm\,0.077$ & $25.172\,\pm\,0.328$ & 52.776\\
Yeast & $7.864\,\pm\,0.131$ & $5.521\,\pm\,0.108$ & $5.413\,\pm\,0.125$ & $25.437\,\pm\,0.362$ & 41.638\\
Glass & $6.768\,\pm\,0.117$ & $5.998\,\pm\,0.109$ & $5.998\,\pm\,0.109$ & $24.011\,\pm\,0.368$ & 43.824\\
Wine & $7.849\,\pm\,0.062$ & $7.255\,\pm\,0.058$ & $7.255\,\pm\,0.058$ & $23.007\,\pm\,0.271$ & 42.668\\
Statlog & $6.519\,\pm\,0.138$ & $9.578\,\pm\,0.132$ & $9.557\,\pm\,0.142$ & $24.846\,\pm\,0.568$ & 48.201\\
WDBC & $8.088\,\pm\,0.066$ & $11.832\,\pm\,0.104$ & $11.832\,\pm\,0.104$ & $21.834\,\pm\,0.202$ & 40.129\\
Ionosphere & $7.241\,\pm\,0.097$ & $13.861\,\pm\,0.146$ & $13.861\,\pm\,0.146$ & $21.803\,\pm\,0.346$ & 37.223\\
Sonar & $8.675\,\pm\,0.113$ & $17.107\,\pm\,0.195$ & $17.107\,\pm\,0.195$ & $21.343\,\pm\,0.216$ & 31.600\\
Worms-64 & $9.064\,\pm\,0.139$ & $13.742\,\pm\,0.173$ & $13.742\,\pm\,0.173$ & $13.352\,\pm\,0.180$ & 16.423\\
MNIST & $10.201\,\pm\,0.115$ & $22.047\,\pm\,0.357$ & $16.465\,\pm\,0.213$ & $18.990\,\pm\,0.267$ & 23.943\\
Fashion-MNIST & $7.814\,\pm\,0.100$ & $29.713\,\pm\,0.541$ & $17.215\,\pm\,0.259$ & $21.080\,\pm\,0.320$ & 34.937\\
\bottomrule
\end{tabular}
\end{table}

\begin{table}[p]
\centering\scriptsize
\setlength{\tabcolsep}{3pt}
\caption{Squared Euclidean excess matching cost at $\eps=4$, $n=64$. Entries are means $\pm$ pointwise 95\% confidence half-widths over twenty parent collections. RAND denotes $\mathrm{RAND}-\mathrm{OPT}$.}
\label{tab:exp-bench-p2-eps4}
\begin{tabular}{lrrrrr}
\toprule
Dataset & Gumbel & Original points & \shortstack{Projected\\representation} & Distance vectors & RAND\\
\midrule
Ring noisy & $59.764\,\pm\,1.953$ & $15.084\,\pm\,0.228$ & $15.084\,\pm\,0.228$ & $87.897\,\pm\,1.756$ & 115.118\\
Circles & $50.292\,\pm\,0.638$ & $14.914\,\pm\,0.192$ & $14.914\,\pm\,0.192$ & $81.599\,\pm\,0.615$ & 120.965\\
Chainlink & $55.844\,\pm\,0.741$ & $22.073\,\pm\,0.294$ & $22.073\,\pm\,0.294$ & $83.648\,\pm\,0.957$ & 118.613\\
Iris & $35.820\,\pm\,0.508$ & $16.634\,\pm\,0.382$ & $16.634\,\pm\,0.382$ & $67.329\,\pm\,0.717$ & 120.007\\
Banknote & $60.454\,\pm\,1.069$ & $27.045\,\pm\,0.482$ & $27.045\,\pm\,0.482$ & $91.231\,\pm\,1.102$ & 115.218\\
Ecoli & $61.596\,\pm\,0.699$ & $41.045\,\pm\,0.411$ & $37.895\,\pm\,0.628$ & $87.604\,\pm\,0.571$ & 108.964\\
Yeast & $63.979\,\pm\,1.013$ & $43.410\,\pm\,0.756$ & $40.992\,\pm\,0.715$ & $80.105\,\pm\,1.993$ & 87.807\\
Glass & $62.465\,\pm\,0.970$ & $43.797\,\pm\,0.585$ & $43.797\,\pm\,0.585$ & $83.081\,\pm\,1.641$ & 90.674\\
Wine & $58.688\,\pm\,0.975$ & $55.376\,\pm\,0.404$ & $55.376\,\pm\,0.404$ & $78.604\,\pm\,0.637$ & 94.353\\
Statlog & $58.330\,\pm\,1.754$ & $59.246\,\pm\,1.472$ & $58.289\,\pm\,1.472$ & $84.449\,\pm\,2.020$ & 95.863\\
WDBC & $57.725\,\pm\,0.916$ & $62.689\,\pm\,0.603$ & $62.689\,\pm\,0.603$ & $79.612\,\pm\,0.792$ & 93.257\\
Ionosphere & $53.854\,\pm\,0.727$ & $58.583\,\pm\,0.828$ & $58.261\,\pm\,0.829$ & $66.316\,\pm\,1.062$ & 74.116\\
Sonar & $53.709\,\pm\,0.419$ & $63.241\,\pm\,0.626$ & $63.241\,\pm\,0.626$ & $66.900\,\pm\,0.599$ & 73.229\\
Worms-64 & $28.295\,\pm\,0.474$ & $28.691\,\pm\,0.484$ & $28.613\,\pm\,0.457$ & $29.777\,\pm\,0.480$ & 30.584\\
MNIST & $47.499\,\pm\,0.546$ & $56.404\,\pm\,0.704$ & $53.069\,\pm\,0.657$ & $54.231\,\pm\,0.663$ & 57.572\\
Fashion-MNIST & $52.339\,\pm\,0.518$ & $77.530\,\pm\,1.195$ & $68.736\,\pm\,0.993$ & $69.436\,\pm\,0.866$ & 80.839\\
\bottomrule
\end{tabular}
\end{table}

\begin{table}[p]
\centering\scriptsize
\setlength{\tabcolsep}{3pt}
\caption{Squared Euclidean excess matching cost at $\eps=8$, $n=64$. Entries are means $\pm$ pointwise 95\% confidence half-widths over twenty parent collections. RAND denotes $\mathrm{RAND}-\mathrm{OPT}$.}
\label{tab:exp-bench-p2-eps8}
\begin{tabular}{lrrrrr}
\toprule
Dataset & Gumbel & Original points & \shortstack{Projected\\representation} & Distance vectors & RAND\\
\midrule
Ring noisy & $34.186\,\pm\,1.955$ & $4.331\,\pm\,0.101$ & $4.331\,\pm\,0.101$ & $70.761\,\pm\,1.915$ & 115.118\\
Circles & $26.482\,\pm\,0.502$ & $4.245\,\pm\,0.076$ & $4.245\,\pm\,0.076$ & $59.780\,\pm\,0.613$ & 120.965\\
Chainlink & $31.528\,\pm\,0.493$ & $6.246\,\pm\,0.074$ & $6.246\,\pm\,0.074$ & $64.207\,\pm\,0.932$ & 118.613\\
Iris & $18.960\,\pm\,0.229$ & $6.069\,\pm\,0.121$ & $6.069\,\pm\,0.121$ & $42.697\,\pm\,0.589$ & 120.007\\
Banknote & $37.000\,\pm\,0.925$ & $9.430\,\pm\,0.210$ & $9.430\,\pm\,0.210$ & $74.550\,\pm\,1.366$ & 115.218\\
Ecoli & $36.535\,\pm\,0.620$ & $16.707\,\pm\,0.203$ & $14.919\,\pm\,0.311$ & $71.737\,\pm\,0.677$ & 108.964\\
Yeast & $46.604\,\pm\,1.413$ & $21.019\,\pm\,0.354$ & $19.303\,\pm\,0.387$ & $73.027\,\pm\,1.509$ & 87.807\\
Glass & $42.019\,\pm\,0.780$ & $19.799\,\pm\,0.423$ & $19.799\,\pm\,0.423$ & $75.813\,\pm\,1.477$ & 90.674\\
Wine & $36.907\,\pm\,0.775$ & $30.284\,\pm\,0.265$ & $30.284\,\pm\,0.265$ & $66.372\,\pm\,0.659$ & 94.353\\
Statlog & $37.526\,\pm\,1.412$ & $32.102\,\pm\,0.845$ & $31.137\,\pm\,0.771$ & $74.668\,\pm\,2.121$ & 95.863\\
WDBC & $38.998\,\pm\,0.933$ & $40.248\,\pm\,0.558$ & $40.248\,\pm\,0.558$ & $69.153\,\pm\,0.833$ & 93.257\\
Ionosphere & $36.477\,\pm\,0.538$ & $43.315\,\pm\,0.563$ & $42.796\,\pm\,0.544$ & $59.296\,\pm\,0.914$ & 74.116\\
Sonar & $37.549\,\pm\,0.427$ & $53.101\,\pm\,0.444$ & $53.101\,\pm\,0.444$ & $61.179\,\pm\,0.500$ & 73.229\\
Worms-64 & $25.775\,\pm\,0.474$ & $26.799\,\pm\,0.465$ & $26.722\,\pm\,0.424$ & $28.979\,\pm\,0.490$ & 30.584\\
MNIST & $35.823\,\pm\,0.377$ & $55.164\,\pm\,0.703$ & $48.262\,\pm\,0.577$ & $50.883\,\pm\,0.598$ & 57.572\\
Fashion-MNIST & $33.151\,\pm\,0.562$ & $74.128\,\pm\,1.098$ & $56.452\,\pm\,0.670$ & $60.051\,\pm\,0.682$ & 80.839\\
\bottomrule
\end{tabular}
\end{table}

\begin{table}[p]
\centering\scriptsize
\setlength{\tabcolsep}{3pt}
\caption{Squared Euclidean excess matching cost at $\eps=16$, $n=64$. Entries are means $\pm$ pointwise 95\% confidence half-widths over twenty parent collections. RAND denotes $\mathrm{RAND}-\mathrm{OPT}$.}
\label{tab:exp-bench-p2-eps16}
\begin{tabular}{lrrrrr}
\toprule
Dataset & Gumbel & Original points & \shortstack{Projected\\representation} & Distance vectors & RAND\\
\midrule
Ring noisy & $15.298\,\pm\,1.107$ & $1.315\,\pm\,0.050$ & $1.315\,\pm\,0.050$ & $52.260\,\pm\,1.837$ & 115.118\\
Circles & $12.953\,\pm\,0.288$ & $1.116\,\pm\,0.021$ & $1.116\,\pm\,0.021$ & $39.429\,\pm\,0.555$ & 120.965\\
Chainlink & $14.216\,\pm\,0.229$ & $1.650\,\pm\,0.034$ & $1.650\,\pm\,0.034$ & $45.803\,\pm\,0.743$ & 118.613\\
Iris & $11.434\,\pm\,0.168$ & $2.354\,\pm\,0.055$ & $2.354\,\pm\,0.055$ & $25.079\,\pm\,0.348$ & 120.007\\
Banknote & $19.961\,\pm\,0.591$ & $2.732\,\pm\,0.050$ & $2.732\,\pm\,0.050$ & $55.296\,\pm\,1.254$ & 115.218\\
Ecoli & $18.809\,\pm\,0.351$ & $5.384\,\pm\,0.079$ & $4.736\,\pm\,0.110$ & $53.107\,\pm\,0.628$ & 108.964\\
Yeast & $27.891\,\pm\,1.467$ & $7.472\,\pm\,0.151$ & $6.738\,\pm\,0.188$ & $62.284\,\pm\,1.412$ & 87.807\\
Glass & $23.282\,\pm\,0.600$ & $6.787\,\pm\,0.255$ & $6.787\,\pm\,0.255$ & $63.690\,\pm\,1.053$ & 90.674\\
Wine & $19.076\,\pm\,0.444$ & $11.831\,\pm\,0.141$ & $11.831\,\pm\,0.141$ & $50.907\,\pm\,0.617$ & 94.353\\
Statlog & $20.573\,\pm\,1.087$ & $11.736\,\pm\,0.252$ & $11.248\,\pm\,0.245$ & $60.454\,\pm\,1.972$ & 95.863\\
WDBC & $23.192\,\pm\,0.711$ & $19.131\,\pm\,0.297$ & $19.131\,\pm\,0.297$ & $55.421\,\pm\,0.998$ & 93.257\\
Ionosphere & $18.465\,\pm\,0.456$ & $22.610\,\pm\,0.312$ & $22.115\,\pm\,0.295$ & $48.746\,\pm\,0.715$ & 74.116\\
Sonar & $20.692\,\pm\,0.324$ & $35.405\,\pm\,0.264$ & $35.405\,\pm\,0.264$ & $52.167\,\pm\,0.441$ & 73.229\\
Worms-64 & $20.569\,\pm\,0.540$ & $23.107\,\pm\,0.397$ & $22.959\,\pm\,0.385$ & $27.487\,\pm\,0.489$ & 30.584\\
MNIST & $18.143\,\pm\,0.299$ & $52.628\,\pm\,0.620$ & $38.077\,\pm\,0.332$ & $45.104\,\pm\,0.556$ & 57.572\\
Fashion-MNIST & $16.168\,\pm\,0.334$ & $67.118\,\pm\,0.942$ & $35.844\,\pm\,0.448$ & $47.529\,\pm\,0.537$ & 80.839\\
\bottomrule
\end{tabular}
\end{table}

\clearpage
{\scriptsize
\setlength{\tabcolsep}{3pt}
\begin{longtable}{rrrrrrr}
\caption{Fashion-MNIST dimension sweep, $n=800$, $\eps=16$. Mechanism entries are total excess-cost means $\pm$ pointwise 95\% confidence half-widths. RAND is $\mathrm{RAND}-\mathrm{OPT}$; $B=2n\min\{d,\log n\}/\eps$.}\label{tab:exp-fashion-values}\\
\toprule
$d$ & Gumbel & Original points & \shortstack{Projected\\representation} & Distance vectors & RAND & $B$\\
\midrule\endfirsthead
\toprule
$d$ & Gumbel & Original points & \shortstack{Projected\\representation} & Distance vectors & RAND & $B$\\
\midrule\endhead
\bottomrule\endfoot
1 & $65.556\,\pm\,1.052$ & $33.914\,\pm\,0.265$ & $33.914\,\pm\,0.265$ & $296.420\,\pm\,2.252$ & 851.812 & 100.000\\
2 & $127.698\,\pm\,0.602$ & $52.767\,\pm\,0.460$ & $52.767\,\pm\,0.460$ & $463.815\,\pm\,4.463$ & 903.867 & 200.000\\
3 & $155.818\,\pm\,0.409$ & $65.883\,\pm\,0.507$ & $65.883\,\pm\,0.507$ & $508.928\,\pm\,5.984$ & 879.018 & 300.000\\
4 & $169.041\,\pm\,0.355$ & $76.365\,\pm\,0.514$ & $76.365\,\pm\,0.514$ & $512.379\,\pm\,5.700$ & 840.827 & 400.000\\
5 & $176.880\,\pm\,0.396$ & $86.390\,\pm\,0.515$ & $86.390\,\pm\,0.515$ & $505.198\,\pm\,5.196$ & 805.214 & 500.000\\
6 & $181.618\,\pm\,0.451$ & $95.989\,\pm\,0.472$ & $95.989\,\pm\,0.472$ & $495.192\,\pm\,4.711$ & 776.106 & 600.000\\
7 & $184.459\,\pm\,0.508$ & $104.865\,\pm\,0.457$ & $104.865\,\pm\,0.457$ & $487.454\,\pm\,4.376$ & 752.415 & 668.461\\
8 & $186.110\,\pm\,0.453$ & $113.312\,\pm\,0.347$ & $113.312\,\pm\,0.347$ & $480.688\,\pm\,3.694$ & 733.505 & 668.461\\
9 & $187.310\,\pm\,0.511$ & $121.565\,\pm\,0.309$ & $121.565\,\pm\,0.309$ & $473.888\,\pm\,3.502$ & 718.097 & 668.461\\
10 & $188.057\,\pm\,0.616$ & $129.248\,\pm\,0.283$ & $129.248\,\pm\,0.283$ & $468.736\,\pm\,2.996$ & 705.000 & 668.461\\
11 & $188.419\,\pm\,0.692$ & $136.556\,\pm\,0.276$ & $136.556\,\pm\,0.276$ & $464.097\,\pm\,3.111$ & 694.101 & 668.461\\
12 & $188.441\,\pm\,0.633$ & $143.597\,\pm\,0.207$ & $143.597\,\pm\,0.207$ & $459.293\,\pm\,2.920$ & 685.759 & 668.461\\
14 & $188.353\,\pm\,0.816$ & $156.747\,\pm\,0.350$ & $156.747\,\pm\,0.350$ & $453.078\,\pm\,2.687$ & 671.547 & 668.461\\
19 & $188.143\,\pm\,0.740$ & $185.864\,\pm\,0.497$ & $185.864\,\pm\,0.497$ & $442.387\,\pm\,2.290$ & 648.225 & 668.461\\
24 & $187.388\,\pm\,0.745$ & $210.512\,\pm\,0.611$ & $210.512\,\pm\,0.611$ & $436.357\,\pm\,1.901$ & 634.520 & 668.461\\
29 & $186.591\,\pm\,0.832$ & $231.696\,\pm\,0.848$ & $231.696\,\pm\,0.848$ & $432.399\,\pm\,1.916$ & 626.210 & 668.461\\
34 & $185.913\,\pm\,0.870$ & $250.188\,\pm\,0.922$ & $250.188\,\pm\,0.922$ & $429.403\,\pm\,1.889$ & 620.069 & 668.461\\
39 & $185.590\,\pm\,0.766$ & $266.560\,\pm\,1.196$ & $266.560\,\pm\,1.196$ & $427.358\,\pm\,2.095$ & 615.051 & 668.461\\
44 & $185.064\,\pm\,0.732$ & $280.933\,\pm\,1.241$ & $280.933\,\pm\,1.241$ & $425.563\,\pm\,1.995$ & 612.022 & 668.461\\
49 & $184.739\,\pm\,0.751$ & $293.727\,\pm\,1.225$ & $293.727\,\pm\,1.225$ & $424.060\,\pm\,1.908$ & 609.377 & 668.461\\
54 & $184.547\,\pm\,0.763$ & $305.562\,\pm\,1.425$ & $305.562\,\pm\,1.425$ & $423.357\,\pm\,1.981$ & 606.543 & 668.461\\
59 & $184.279\,\pm\,0.825$ & $315.995\,\pm\,1.479$ & $315.995\,\pm\,1.479$ & $422.279\,\pm\,1.983$ & 604.979 & 668.461\\
64 & $184.048\,\pm\,0.794$ & $325.447\,\pm\,1.478$ & $325.447\,\pm\,1.478$ & $421.351\,\pm\,1.985$ & 603.348 & 668.461\\
69 & $183.764\,\pm\,0.752$ & $334.038\,\pm\,1.466$ & $334.038\,\pm\,1.466$ & $420.633\,\pm\,1.927$ & 602.115 & 668.461\\
74 & $183.635\,\pm\,0.782$ & $342.051\,\pm\,1.604$ & $342.051\,\pm\,1.604$ & $420.124\,\pm\,1.936$ & 600.603 & 668.461\\
79 & $183.492\,\pm\,0.742$ & $349.199\,\pm\,1.596$ & $349.199\,\pm\,1.596$ & $419.541\,\pm\,1.872$ & 599.486 & 668.461\\
84 & $183.524\,\pm\,0.744$ & $355.947\,\pm\,1.549$ & $355.947\,\pm\,1.549$ & $419.042\,\pm\,1.739$ & 598.232 & 668.461\\
89 & $183.381\,\pm\,0.696$ & $362.184\,\pm\,1.477$ & $362.184\,\pm\,1.477$ & $418.421\,\pm\,1.707$ & 597.376 & 668.461\\
94 & $183.283\,\pm\,0.681$ & $367.951\,\pm\,1.441$ & $367.951\,\pm\,1.441$ & $418.242\,\pm\,1.622$ & 596.566 & 668.461\\
99 & $183.187\,\pm\,0.610$ & $373.384\,\pm\,1.471$ & $373.384\,\pm\,1.471$ & $417.901\,\pm\,1.630$ & 595.939 & 668.461\\
104 & $183.079\,\pm\,0.591$ & $378.428\,\pm\,1.496$ & $378.428\,\pm\,1.496$ & $417.639\,\pm\,1.578$ & 595.442 & 668.461\\
109 & $182.920\,\pm\,0.598$ & $383.089\,\pm\,1.527$ & $383.089\,\pm\,1.527$ & $417.264\,\pm\,1.613$ & 595.207 & 668.461\\
114 & $182.786\,\pm\,0.574$ & $387.459\,\pm\,1.458$ & $387.459\,\pm\,1.458$ & $417.006\,\pm\,1.619$ & 594.677 & 668.461\\
119 & $182.802\,\pm\,0.541$ & $391.677\,\pm\,1.452$ & $391.677\,\pm\,1.452$ & $416.616\,\pm\,1.579$ & 594.234 & 668.461\\
124 & $182.602\,\pm\,0.565$ & $395.535\,\pm\,1.494$ & $395.535\,\pm\,1.494$ & $416.539\,\pm\,1.561$ & 594.072 & 668.461\\
129 & $182.522\,\pm\,0.508$ & $399.354\,\pm\,1.425$ & $399.354\,\pm\,1.425$ & $416.342\,\pm\,1.539$ & 593.683 & 668.461\\
134 & $182.535\,\pm\,0.538$ & $402.976\,\pm\,1.475$ & $402.976\,\pm\,1.475$ & $416.251\,\pm\,1.527$ & 593.084 & 668.461\\
139 & $182.512\,\pm\,0.587$ & $406.063\,\pm\,1.485$ & $406.063\,\pm\,1.485$ & $415.989\,\pm\,1.519$ & 592.625 & 668.461\\
144 & $182.478\,\pm\,0.604$ & $409.245\,\pm\,1.507$ & $409.245\,\pm\,1.507$ & $415.778\,\pm\,1.509$ & 592.258 & 668.461\\
149 & $182.440\,\pm\,0.558$ & $412.341\,\pm\,1.523$ & $412.341\,\pm\,1.523$ & $415.706\,\pm\,1.489$ & 591.982 & 668.461\\
154 & $182.393\,\pm\,0.599$ & $415.240\,\pm\,1.489$ & $415.240\,\pm\,1.489$ & $415.536\,\pm\,1.496$ & 591.722 & 668.461\\
159 & $182.371\,\pm\,0.574$ & $417.923\,\pm\,1.531$ & $417.923\,\pm\,1.531$ & $415.415\,\pm\,1.490$ & 591.492 & 668.461\\
164 & $182.332\,\pm\,0.590$ & $420.528\,\pm\,1.565$ & $420.528\,\pm\,1.565$ & $415.221\,\pm\,1.520$ & 591.338 & 668.461\\
169 & $182.310\,\pm\,0.591$ & $423.090\,\pm\,1.553$ & $423.090\,\pm\,1.553$ & $415.236\,\pm\,1.500$ & 591.146 & 668.461\\
174 & $182.228\,\pm\,0.594$ & $425.420\,\pm\,1.574$ & $425.420\,\pm\,1.574$ & $415.100\,\pm\,1.482$ & 590.818 & 668.461\\
179 & $182.198\,\pm\,0.569$ & $427.758\,\pm\,1.593$ & $427.758\,\pm\,1.593$ & $415.075\,\pm\,1.497$ & 590.577 & 668.461\\
184 & $182.156\,\pm\,0.548$ & $430.035\,\pm\,1.586$ & $430.035\,\pm\,1.586$ & $414.976\,\pm\,1.473$ & 590.417 & 668.461\\
189 & $182.107\,\pm\,0.546$ & $432.279\,\pm\,1.565$ & $432.279\,\pm\,1.565$ & $414.917\,\pm\,1.497$ & 590.467 & 668.461\\
194 & $181.998\,\pm\,0.553$ & $434.260\,\pm\,1.623$ & $434.260\,\pm\,1.623$ & $414.853\,\pm\,1.574$ & 590.349 & 668.461\\
199 & $182.029\,\pm\,0.554$ & $436.081\,\pm\,1.634$ & $436.081\,\pm\,1.634$ & $414.670\,\pm\,1.537$ & 590.111 & 668.461\\
204 & $181.981\,\pm\,0.581$ & $438.166\,\pm\,1.619$ & $438.166\,\pm\,1.619$ & $414.668\,\pm\,1.507$ & 589.868 & 668.461\\
254 & $181.868\,\pm\,0.601$ & $453.813\,\pm\,1.651$ & $453.813\,\pm\,1.651$ & $414.083\,\pm\,1.457$ & 588.641 & 668.461\\
304 & $181.773\,\pm\,0.577$ & $465.543\,\pm\,1.652$ & $465.543\,\pm\,1.652$ & $413.482\,\pm\,1.483$ & 587.824 & 668.461\\
404 & $181.573\,\pm\,0.549$ & $481.966\,\pm\,1.629$ & $481.966\,\pm\,1.629$ & $413.179\,\pm\,1.458$ & 586.724 & 668.461\\
604 & $181.446\,\pm\,0.565$ & $501.005\,\pm\,1.696$ & $501.005\,\pm\,1.696$ & $412.602\,\pm\,1.480$ & 585.394 & 668.461\\
784 & $181.312\,\pm\,0.562$ & $511.643\,\pm\,1.637$ & $511.655\,\pm\,1.643$ & $412.208\,\pm\,1.400$ & 584.989 & 668.461\\
\end{longtable}
}